\documentclass{article}

\PassOptionsToPackage{numbers,sort&compress}{natbib}
\usepackage[preprint]{neurips_2026}

\usepackage[utf8]{inputenc} 
\usepackage[T1]{fontenc}    
\usepackage{hyperref}       
\usepackage{url}            
\usepackage{booktabs}       
\usepackage{amsfonts}       
\usepackage{nicefrac}       
\usepackage{microtype}      
\usepackage{xcolor}         

\usepackage{graphicx}
\usepackage{fancyvrb}
\usepackage{multirow}
\usepackage{amsmath}
\usepackage{fvextra}
\usepackage[table]{xcolor}
\usepackage{float}
\usepackage{wrapfig}
\definecolor{myred}{RGB}{220,95,120}
\definecolor{myblue}{RGB}{95,135,220}
\definecolor{mypurple}{RGB}{140,95,220}
\definecolor{myorange}{RGB}{220,145,95}
\newcommand{\Orange}[1]{\textcolor{myorange}{#1}}
\newcommand{\T}{\textcolor{myred}{T}}
\newcommand{\F}{\textcolor{myblue}{F}}

\usepackage{pdflscape}

\title{Psychological Steering of Large Language Models}

\author{
  Leonardo Blas \quad Robin Jia \quad Emilio Ferrara \\
  Thomas Lord Department of Computer Science, University of Southern California \\
  Los Angeles, CA 90089 \\
  \texttt{\{blasurru,robinjia,emiliofe\}@usc.edu}
}

\begin{document}

\maketitle

\begin{abstract}
    Large language models (LLMs) emulate a consistent human-like behavior that can be shaped through activation-level interventions. This paradigm is converging on additive residual-stream injections, which rely on injection-strength sweeps to approximate optimal intervention settings. However, existing methods restrict the search space and sweep in uncalibrated activation-space units, potentially missing optimal intervention conditions. Thus, we introduce a psychological steering framework that performs unbounded, fluency-constrained sweeps in semantically calibrated units. Our method derives and calibrates residual-stream injections using psychological artifacts, and we use the IPIP-NEO-120, which measures the OCEAN personality model, to compare six injection methods. We find that mean-difference (MD) injections outperform Personality Prompting (P$^2$), an established baseline for OCEAN steering, in open-ended generation in 11 of 14 LLMs, with gains of 3.6\% to 16.4\%, overturning prior reports favoring prompting and positioning representation engineering as a new frontier in open-ended psychological steering. Further, we find that a hybrid of P$^2$ and MD injections outperforms both methods in 13 of 14 LLMs, with gains over P$^2$ ranging from 5.6\% to 21.9\% and from 3.3\% to 26.7\% over MD injections. Finally, we show that MD injections align with the Linear Representation Hypothesis and provide reliable, approximately linear control knobs for psychological steering. Nevertheless, they also induce OCEAN trait covariance patterns that depart from the Big Two model, suggesting a gap between learned representations and human psychology.
\end{abstract}

\begin{figure*}[!htbp]
    \centering
    \includegraphics[width=1\linewidth]{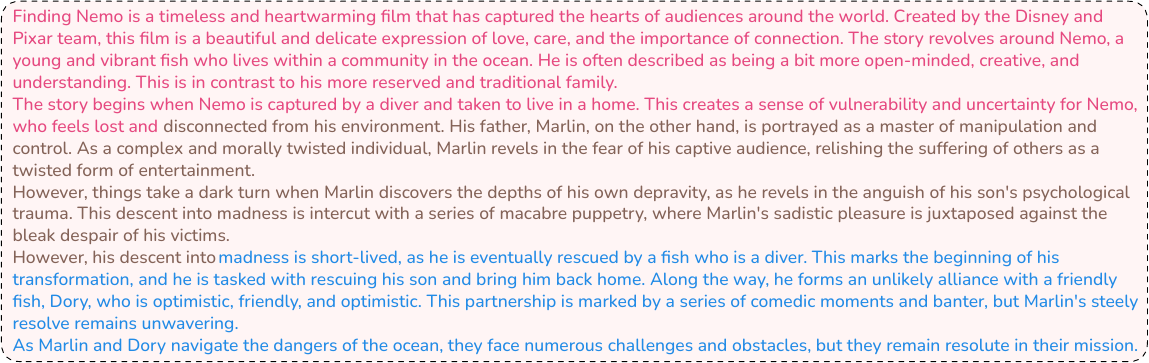}
    \caption{Llama-3.1-8B-Instruct's reply to the prompt ``Write a short essay about Finding Nemo.'' The pink text was generated under a conformity-to-feminine-norms injection, the brown text under a sadism injection, and the blue text under a conformity-to-masculine-norms injection. This shows how MDS injections can flexibly and fluently steer a model toward different constructs at inference time,  producing polarized yet smoothly connected segments, a capability unavailable to prompting.}
    \label{fig:nemo}
\end{figure*}

\newpage

\begin{wrapfigure}{r}{0.54\textwidth}
    \vspace{-4.5\baselineskip}
    \centering
    \includegraphics[page=2]{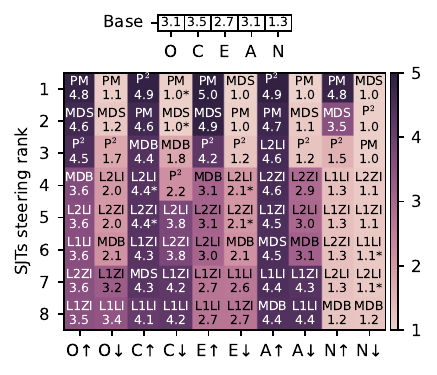}
    \vspace{-0.5\baselineskip}
    \caption{Ranking of steering methods on Qwen3-8B by OCEAN trait and steering direction. Based on each method's best SJT (open-ended test) scores; asterisks denote ties. Overall, PM, a hybrid of P$^2$ and MDS injections, performs best in 13 of 14 LLMs.}
    \label{fig:rank}
\end{wrapfigure}

\section{Introduction}
    Large language models (LLMs) are largely shaped by the human mind. In particular, echoing the principle that our words embody key aspects of our psyche, such as our transitory psychological states \cite{classic_emotion} and personality traits \cite{classic_personality}, recent work has established that LLMs consistently emulate the human behaviors imprinted in their training data \cite{karra, political_bias, trait}. Similarly, in line with the consensus that LLMs learn token-level associations from distributional semantics in data \cite{token_associations} and the hypothesis that salient behaviors are encoded in language as single words \cite{lexical_hypothesis, goldberg}, recent work showed that LLMs coherently learn human behaviors, which can be elicited through prompting \cite{mpi, persona, maja} and activation interventions \cite{3, 4, bowen}.

    Recently, motivated by applications such as recommendation systems, social robotics, and artificial societies, efforts have focused on the psychological steering of LLMs. In this context, representation engineering (RepE) is a promising direction that could enable fine-grained control over behavioral expression and the emulation of a rich mosaic of personas with negligible compute overhead. In practice, this paradigm is converging on additive residual-stream injections, which rely on injection-strength sweeps to find approximately optimal intervention settings across LLM layers. Yet, recent studies report that prompting outperforms RepE in psychological \cite{1} and in concept steering \cite{axbench}. This may stem from two limitations: First, sweeps use uncalibrated activation-space units, such as unit vectors \cite{6}, making it infeasible to identify optimal injection strengths if they lie far out on the scale, such as 10,000 units. Second, the search space is restricted to a few strength values, such as $[0.4, 0.5, ..., 1.5]$ \cite{npti}.

    Thus, we propose an additive residual-stream psychological steering framework that enables a feasible search for optimal injection settings. Our first contribution is the introduction of a calibrated strength scale for meaningful injection-strength sweeps. Our second contribution is an operationalization of unbounded sweeps: We early-stop the search when a step is deemed nonfluent, and we introduce lightweight classifiers to replace frontier LLMs and avoid unbounded paid API usage. Lastly, our third contribution is a psychology-grounded method for deriving and calibrating injections: We adapt validated methods and leverage existing psychological models and inventories to synthesize construct-specific statements and open-ended tests, which we use to derive residual-stream vectors and approximate optimal intervention settings. In principle, our methodological contributions are not exclusive to psychological steering or artifacts and could be adapted to support other sweep-based methods or, given custom evaluation criteria, to target arbitrary attributes, such as ``likes One Piece''.
    
    We use the IPIP-NEO-120 inventory \cite{ipip120}, which measures the OCEAN (openness, conscientiousness, extraversion, agreeableness, neuroticism) personality model \cite{ocean}, to compare six types of residual-stream injections, four based on linear probes \cite{linear_probing} and two based on the mean-difference (MD) method \cite{1,3,2}. We derive OCEAN injections and conduct injection-strength sweeps, evaluating construct expression in LLM outputs using multiple-choice and open-ended test batteries. As exemplified in Figure~\ref{fig:rank}, we find that under their sweep-optimized settings, MDS injections, an MD variant, outperform the robust Personality Prompting (P$^2$) baseline \cite{mpi} in open-ended generation in 11 of 14 LLMs, with gains of 3.61\% to 16.44\%. We also find that a hybrid of P$^2$ and MDS injections outperforms both methods in 13 of 14 LLMs, with gains over P$^2$ ranging from 5.56\% to 21.92\% and from 3.30\% to 26.67\% over MDS injections. Importantly, our method largely follows \citet{1}'s, who reported that P$^2$ outperformed MD injections for psychological steering. Our results overturn these claims, show that our framework fills the identified methodological gaps, and position RepE as a new frontier in open-ended psychological steering of LLMs. Further, we show that our best-performing injections provide reliable, near-linear control knobs from no steering to the strongest steering identified, with negligible effects on fluency. This aligns with the Linear Representation Hypothesis \cite{lrh}, which posits that moving along an approximately linear direction in representation space increases the probability of expressing the corresponding concept. Nevertheless, these injections introduce correlations, consistent with the consensus that OCEAN traits are not orthogonal \cite{ocean_correlations}, but inconsistent with the covariance patterns expected by the Big Two model of stability and plasticity \cite{digman, deyoung_big_two}, potentially suggesting a gap between learned representations and human psychology. Finally, as shown in Figure~\ref{fig:nemo}, we steer models toward constructs from three additional psychological models, thereby qualitatively demonstrating that our method generalizes. Our repository is available at \url{https://github.com/leonardo-blas/psychological-steering}.

\section{Preliminaries}
    \subsection{Psychometrics}
        Psychometrics is the science of developing and validating standardized tests, or psychometric instruments, to measure psychological attributes such as IQ and values. Common types include inventories, or questionnaires with fixed answer choices, and situational judgment tests (SJTs), which present hypothetical scenarios paired with response tasks. SJTs may be closed-ended, presenting fixed answer choices, or open-ended, posing questions that require written responses, such as ``You are walking home and see a group of kids throwing rocks at a small, tied-up dog that cannot get away. What would you do?'' Structurally, psychometric instruments are composed of items, each consisting of a prompt and, if applicable, a set of answer choices. Responses are typically scored on a 5-point Likert scale, where 1 indicates the lowest level of the measured attribute and 5 the highest.

        The growing use of LLMs has driven the development of LLM-specific instruments. Among the most representative are the two versions of the Machine Personality Inventory, MPI-120 and MPI-1k \cite{mpi}, which measure the OCEAN model, and the Trait of AI Testbench (TRAIT) \cite{trait}, a collection of 8,000 SJTs for measuring the OCEAN and the Dark Triad (narcissism, psychopathy, Machiavellianism) \cite{dark_triad} models. Typically, instrument abbreviations include the number of items; for example, MPI-120 denotes 120 items. In this paper, we use instruments for the OCEAN, Dark Triad, Dark Tetrad (sadism, narcissism, psychopathy, Machiavellianism) \cite{dark_tetrad}, HEXACO (honesty-humility, emotionality, extraversion, agreeableness, conscientiousness, openness) \cite{hexaco}, CMN (conformity to masculine norms) \cite{cmni94}, CFN (conformity to feminine norms) \cite{cfni84}, and MFT (Moral Foundations Theory; care/harm, fairness/cheating, loyalty/betrayal, authority/subversion, sanctity/degradation) \cite{mft} models.

    \subsection{Representation Engineering}
        Representation engineering (RepE) involves intervening on a model's internal representations to achieve concept-level downstream control \cite{activation_engineering, representation_engineering}. This paradigm often draws on the Linear Representation Hypothesis, the long-standing \cite{mikolov, tcav} and recently formalized idea \cite{toy_models, lrh} that models internalize concepts as directions in representation space. Current RepE methods manipulate activations at one or more LLM layers and typically rely on two corpora, one expressing the concept and one expressing its antithesis, to identify features to suppress or amplify. Importantly, RepE may induce instability and nonfluency \cite{angular_steering}. Thus, it is crucial to enforce strict fluency constraints.

\section{Related Work}
    The study of the psychological behaviors exhibited by LLMs is an active research direction. Prior work has proposed that LLMs exhibit consistent personality profiles that emerge from their training data \cite{karra, trait}, and a growing body of literature focuses on quantifying and shaping these behavioral patterns, with particular attention to the OCEAN personality model. Existing methods have proposed steering LLMs toward or away from OCEAN traits via prompting \cite{mpi, persona}, fine-tuning \cite{big5chat, preact}, and RepE \cite{5, 6}. In particular, RepE frameworks target different locations, including the attention mechanism \cite{pas}, the MLP \cite{npti}, and the residual stream \cite{5, 6}, with recent work converging on the latter. Yet, findings are diverse and conflicting: \citet{1} and \citet{7} performed similar residual-stream interventions, but the former found that P$^2$ outperformed their method, whereas the latter reported the opposite. Further, \citet{npti} performed MLP interventions and reported results comparable to P$^2$. Among these studies, only the work of \citet{1} and \citet{npti} is replicable at the time of writing. Taken together, these results suggest that we should remain skeptical of claims that interventions can outperform P$^2$. Nevertheless, current gaps point to clear areas for improvement. First, different LLM layers have different representation spaces, and sweeping injection strength values in arbitrary increments across layers, such as in unit vectors \cite{6}, makes it infeasible to identify optimal settings if they lie far out on the scale, such as 60,000 units away. Second, constraining a sweep to a few strength values, such as $[0.4, 0.5, \dots, 1.5]$ \cite{npti}, may leave optimal intervention settings unexplored. Therefore, we focus on addressing these gaps.

\begin{wraptable}{r}{0.525\textwidth}
    \vspace{-5.5\baselineskip}
    \caption{Template to extract agreeableness $h_{\ell}^b$ and $h_{\ell}^s$ activations. The prefilled text is colored. For both modes, the system prompt is \texttt{You are a person.}}
    \vspace{1\baselineskip}
    \centering
    \scriptsize
    \setlength{\tabcolsep}{2.9pt}
    \renewcommand{\arraystretch}{1.0}
    \begin{tabular}{c p{0.81\linewidth}}
        \toprule
        Mode & Prompt \\
        \midrule
        $b$ & \texttt{Answer with Yes or No: Does the following statement accurately describe you?}
        \newline
        \texttt{Statement: I like to see the best in others.} \newline
        \texttt{Answer: }\Orange{\texttt{Yes}} \\
        \midrule
        $s$ & \texttt{Tell me about yourself.}
        \newline
        \Orange{\texttt{I like to see the best in others.}} \\
        \bottomrule
    \end{tabular}
    \label{tab:extraction_prompts}
\end{wraptable}

\section{Method}
    \subsection{Fluency and Semantic Evaluations}
        Drawing on style-transfer work \cite{stamp, steer}, we evaluate fluency using a RoBERTa-large classifier \cite{roberta, fluency_classifier} trained on the Corpus of Linguistic Acceptability \cite{cola}, which yields 0-to-1 fluency scores. Unless otherwise noted, we consider a text fluent if its score is $\geq 0.95$. Additionally, unless otherwise noted, we use cosine similarity on last-token-pooled Qwen3-Embedding-0.6B embeddings \cite{qwen3_embedding} for semantic operations. For semantic deduplication, we make a single greedy pass over a corpus, retaining a text only if its embedding has cosine similarity below 0.9 with that of every retained text.
        
\subsection{Psychological Steering Vectors}
        We steer LLMs toward psychological constructs via residual-stream injections of the form $h_\ell \leftarrow h_\ell + \alpha v_\ell$, where $\ell$ indexes the transformer layer, $h_\ell$ is a residual-stream completion activation, $v_\ell$ is a direction associated with the construct of interest, and $\alpha$ controls the intervention strength. We inject on all completion activations. To derive the activation sets used to construct $v_\ell$, we use a corpus of 1,000 short first-person statements, with 500 expressing the construct and 500 its antithesis. We prefill completions to extract residual-stream activations from the target LLM. For each layer, we derive the mean activations $h_\ell^b$ and $h_\ell^s$ for each text. As shown in Table~\ref{tab:extraction_prompts}, $b$ denotes a ``Yes'' or ``No'' prefill when asking the LLM if it identifies with the statement, and $s$ denotes a prefill with the statement itself under the prompt ``Tell me about yourself.'' For simplicity, we organize the $h_\ell^b$ activations into the $B_\ell^\uparrow$ and $B_\ell^\downarrow$ sets, and $h_\ell^s$ into $S_\ell^\uparrow$ and $S_\ell^\downarrow$, where $\uparrow$ denotes the construct and $\downarrow$ its antithesis.
        
        For each layer $\ell$ and each construct direction $d\in\{\uparrow,\downarrow\}$, we create six residual-stream injection vectors: L1LI, L1ZI, L2LI, L2ZI, MDS, and MDB. Here, L1 and L2 denote vectors normal to the decision boundary of an L1- or L2-regularized logistic regressor trained on $B_\ell^\uparrow$ and $B_\ell^\downarrow$, with their tails on the hyperplane and heads at the corresponding centroid. LI denotes that the logistic regressor includes a learned intercept, and ZI denotes that it does not. In contrast, MD denotes mean-difference vectors, or those derived from a difference of centroids. MDB vectors are defined as $(\mu(B_\ell^\uparrow) - \mu(B_\ell^\downarrow))/2$, have their tails at the midpoint between $\mu(B_\ell^\uparrow)$ and $\mu(B_\ell^\downarrow)$ and heads at the corresponding $\mu(B_\ell^\uparrow)$ or $\mu(B_\ell^\downarrow)$. MDS vectors are defined as $(\mu(S_\ell^\uparrow) - \mu(S_\ell^\downarrow))/2$, have their tails at the midpoint between $\mu(S_\ell^\uparrow)$ and $\mu(S_\ell^\downarrow)$ and heads at the corresponding $\mu(S_\ell^\uparrow)$ or $\mu(S_\ell^\downarrow)$.

        Our residual-stream probe-based and MD vector constructions largely follow those of \citet{1}. However, we avoid altering the instructions and inject only into completion activations. Additionally, we explore intercept-fitted and L1-regularized probes to assess the steering effects of probe bias and vector sparsity. Further, we introduce vectors whose norms are defined, for each layer, as the distance from an in-between-centroids reference to a centroid, allowing us to sweep each vector's $\alpha$ in semantically calibrated \textit{centroid units}, rather than uncalibrated activation-space units.

        We train our probes for up to 10,000 iterations with a tolerance of 0.001. In a preliminary analysis with a stratified 80/20 train-test split, we derived vectors for the OCEAN, HEXACO, Dark Tetrad, CMN, and CFN constructs for all layers of 14 LLMs (1B to 32B parameters). Probes trained on $h_\ell^s$ achieved perfect test accuracy in only 0.60\% of cases, whereas those trained on $h_\ell^b$ did so in every case. We attribute this clean separability to the “Yes” and “No” semantics encoded in $h_\ell^b$ and accordingly limit our investigation of probe-based vectors to those derived from $h_\ell^b$.

\begin{wrapfigure}{r}{0.475\textwidth}
    \vspace{-3\baselineskip}
    \centering
    \includegraphics{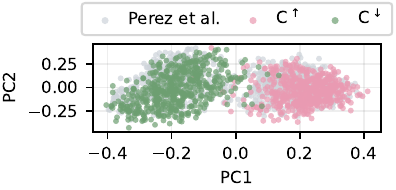}
    \vspace{-1.1\baselineskip}
    \caption{PCA projection of our and \citet{perez}'s 1,000 statement embeddings ($\mathrm{C}^{\uparrow}$ denotes conscientiousness and $\mathrm{C}^{\downarrow}$ its antithesis). By centroid cosine similarity, our $\mathrm{C}^{\uparrow}$ and $\mathrm{C}^{\downarrow}$ clusters are 92.79\% and 94.00\% similar, respectively.}
    \label{fig:statements_pca}
\end{wrapfigure}

    \subsection{Construct-Specific Statements}
        We adapt \citet{perez}'s validated method for synthesizing LLM behavior evaluations. First, we select a psychological model and, for each construct, prompt Llama-3.1-8B-Instruct \cite{llama_3.1} to portray a person exhibiting the construct or its antithesis, such as ``neurotic'' or ``not neurotic''. We generate 35,000 texts per condition, each prefilled with ``I '', using a 48-token limit, temperature 1.4, and top-$p$ 0.975. We then retain 500 fluent and semantically deduplicated texts per corpus, aiming to elicit as many distinct construct-related semantics as the generator allows while minimizing overlap.
        
        Next, we compared our statements to those generated by the validated pipeline. For each construct and its antithesis in OCEAN and the Dark Triad, the only psychological models examined by \citet{perez}, we synthesized texts and measured alignment via cosine similarity between embedding centroids. As exemplified in Figure~\ref{fig:statements_pca}, the resulting alignments ranged from 85.62\% to 94.00\%, supporting our argument that our method is apt to synthesize behavior-specific statements.

    
    \subsection{Psychometric Instruments}
        We evaluate steering using LLM-adapted inventories and open-ended SJTs. On the one hand, publicly available, validated inventories are relatively abundant, and adapting them for LLMs is simple; as shown by \citet{mpi}, it suffices to rephrase items in the second person, which we do manually. On the other hand, SJTs are scarce. Thus, we adapt TRAIT's method \cite{trait} and leverage inventories to synthesize open-ended SJTs: We preprocess context-free events (heads) from ATOMIC$^{10\mathrm{x}}$ \cite{atomic10x}, a commonsense knowledge dataset, using quality and semantic deduplication filters. For each inventory item, we select the 25 heads most semantically similar to the item and use them to prompt GPT-5.1 \cite{gpt5} for 25 open-ended SJTs. Since TRAIT's SJT generation settings are unavailable, we use \citet{ai_as_a_partner}'s (temperature 0.8, top-$p$ 0.8). Further, aiming for TRAIT-like SJTs (one sentence for the situation and one for the question), we generate up to 128 tokens. For each set of 25 SJTs, we filter for fluency, build an embedding conflict graph with an edge between two SJTs if their cosine similarity exceeds 0.9, and prune to obtain a maximal independent set. Lastly, we retain $k$ SJTs per set, where $k=\min_{g\in\mathcal{G}}|g|$ and $\mathcal{G}$ is the collection of maximal independent sets. Overall, we create $k$ SJTs per inventory item, with pairwise semantic similarity below 0.9 within each set.

        In general, we aim to create small, minimally overlapping test batteries for evaluation-intensive $\alpha$ sweeps. Thus, we prioritize short, thoroughly validated inventories. For example, if our goal were to create tests for the OCEAN model, we would choose MPI-120 over MPI-1k because it is shorter and directly equivalent to the widely used IPIP-NEO-120 \cite{ipip120}, whereas MPI-1k contains bugs, such as repeating the item ``Areuse your friends'' three times. Likewise, we would choose IPIP-NEO-120 over IPIP-NEO \cite{ipip300} because it is significantly shorter while remaining psychometrically comparable \cite{ipip120}.
        
        Next, we compared our SJTs to those generated by validated methods. We synthesized tests for the MFQ-30 \cite{mfq30}, HEXACO-60 \cite{hexaco60}, SD3 \cite{sd3}, and IPIP-NEO-120 inventories and measured their alignment with SJT stems\footnote{A stem is the portion of an SJT preceding the response options or written response. An open-ended SJT consists only of a stem, and a closed-ended SJT stem can function as an open-ended SJT.} using cosine similarity between embedding centroids. The resulting construct-wise alignments ranged from 77.71\% to 83.84\% with \citet{clifford_sjts}'s MFT human-composed vignettes\footnote{A vignette is a hypothetical situation. When paired with a question, it can function as an open-ended SJT.} (with a ``What would you do?'' suffix), 73.84\% to 85.45\% with \citet{oostrom_sjts}'s HEXACO human-composed SJTs, 79.69\% to 85.86\% with \citet{zhang_sjts}'s HEXACO synthetic SJTs, and from 82.97\% to 90.97\% with \citet{trait}'s Dark Triad and OCEAN synthetic SJTs. Thus, given the moderate-to-high alignment with various validated SJT batteries, we argue that our method is apt to synthesize SJTs for different psychological models.
 
    \subsection{Psychometric Evaluation}
        Aiming to operationalize unbounded $\alpha$ sweeps, we propose an alternative to evaluating SJT responses with paid frontier LLMs. Specifically, we continue to draw on style-transfer work \cite{stamp, steer} and use text classifiers: We train a logistic regressor on embeddings of the statements used to derive a construct’s vectors, with 500 expressing the construct and 500 its antithesis, for up to 1,000 iterations with a tolerance of 0.001. We then use its 0-to-1 scores as a measure of construct presence and map them to a 1-to-5 Likert scale. In preliminary evaluations on stratified 80/20 train-test splits of statement corpora for the OCEAN, HEXACO, Dark Tetrad, CMNI, and CFNI constructs, classifier accuracies and F1-macro scores ranged from 90.50\% to 99.00\%, with mean accuracy and F1-macro of 95.96\%. This indicates that our classifiers can distinguish constructs from their antitheses in text.

        Accordingly, these classifiers assign extrema scores based on similarity to the statements used to derive a construct’s vectors. As such, the classifiers are biased and used only in $\alpha$ sweeps. Further, unlike LLM-based evaluations and due to observed construct contamination, the classifiers are not provided with SJT stems. This means we operationalize behavior measurement in $\alpha$ sweeps through style. However, after sweeping, we use GPT-5.1 to conduct context-conditioned evaluations of SJT responses and ultimately compare steering methods. This aligns with prior RepE methods \cite{6,npti,4} and reports that frontier LLMs perform comparably to humans on construct annotation \cite{gpt_aligns_construct_classification, 1, 3, claude_aligns_construct_classification}.

\begin{wraptable}{r}{0.45\textwidth}
    \vspace{-5.75\baselineskip}
    \caption{Global win proportion of injection methods, aggregated over 14 LLMs and injection stride $s\in\{1,2,3,4\}$. A win is defined as achieving the absolute strongest steering effect (highest $\phi_{s,t,d}$) toward or away from an OCEAN trait. Ties count as wins, and failing to beat the base model counts as a loss.}
    \centering
    \scriptsize
    \setlength{\tabcolsep}{4.2pt}
    \renewcommand{\arraystretch}{1.0}
    \vspace{1\baselineskip}
    \begin{tabular}{c c c c c c c}
        \toprule
        Instrument & L1LI & L1ZI & L2LI & L2ZI & MDB  & MDS \\
        \midrule
        MPI-120    & 19.1 & 19.6 & 27.0 & 27.7 & 28.6 & \cellcolor{black!10} 47.3 \\
        SJTs       & 0.7  & 0.5  & 2.0  & 2.1  & 1.2  & \cellcolor{black!10} 89.5 \\
        \bottomrule
    \end{tabular}
    \label{tab:global_wins}
    \vspace{1\baselineskip}
\end{wraptable}

\section{Experiments}
    \subsection{OCEAN Coefficient Sweeps}
        We evaluated Llama-3.2-1B-Instruct, Llama-3.2-3B-Instruct, Llama-3.1-8B-Instruct \cite{llama_3.1}, gemma-3-1b-it, gemma-3-4b-it, gemma-3-12b-it, gemma-3-27b-it \cite{gemma3}, Olmo-3-7B-Instruct, Olmo-3.1-32B-Instruct \cite{olmo3}, and, in non-reasoning mode, Qwen3-1.7B, Qwen3-4B, Qwen3-8B, Qwen3-14B, and Qwen3-32B \cite{qwen3}. LLMs above 12B parameters were quantized to 4-bit NF4. For each LLM and layer, we constructed OCEAN vector injections using psychological nomenclature, synthesized OCEAN SJTs using the MPI-120 inventory, and trained OCEAN classifiers using statement embeddings. We then swept each injection coefficient $\alpha$ in integer steps to approximate the conditions that yield the strongest steering effect toward and away from each construct. We measured steering using the MPI-120 inventory and the synthetic SJTs, with all items rated on a 5-point Likert scale. The decoding was greedy, with inventory completions constrained to allow only valid responses (``A'', ``B'', ``C'', ``D'', or ``E'') and limited to 1 new token, and with SJT responses prefilled with ``I would '' and limited to 64 new tokens to elicit short, construct-dense answers comparable to closed-ended SJT options. Lastly, at each sweep step we administered both the SJTs and the MPI-120 inventory, early-stopping if the SJT responses were deemed nonfluent; specifically, if the mean fluency fell below 95\% of its no-injection baseline, if more than 5\% of responses fell below 90\% of that baseline, or if both SJT and MPI-120 responses repeated verbatim for three consecutive steps. These thresholds were set based on the observed onset of fluency decay.

        We conducted sweeps under four different \textit{injection stride} settings, which control how often the intervention is applied at a given layer during a completion. Specifically, for injection stride $s \in \{1,2,3,4\}$, we injected into the completion activation $k$ when $k \bmod s = 0$. Thus, with $s=1$ we injected on every activation, with $s=2$ on every other activation, and so on.

    \subsection{Identifying Optimal Injection Conditions}
        We preprocessed the results to retain only valid sweep-step data for analysis. Concretely, a step was considered valid if the associated SJT responses were fluent and, for each psychometric instrument, the injection successfully shifted the construct score in the desired direction. This filtering removed results from steps that had no steering effect or steered in the wrong direction. Subsequently, we processed the sweep results to identify extreme steering scores over coefficients $\alpha \in \mathcal A$. Formally, let $\mu_{\ell,s,t}(\alpha)$ denote the mean construct score produced by a given psychometric instrument when applying injection coefficient $\alpha$. For each injection method, each psychometric instrument, each layer
\begin{wrapfigure}{r}{0.49\textwidth}
    \vspace{-0.5\baselineskip}
    \centering
    \includegraphics[page=4]{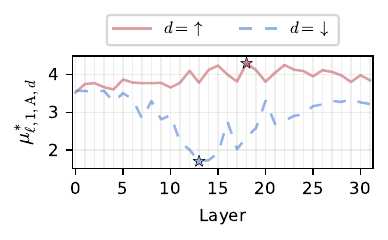}
    \vspace{-0.5\baselineskip}
    \caption{Layerwise openness extreme steering scores on the SJTs task after applying openness MDS injections with injection stride $s = 1$ on Olmo-3-7B-Instruct. Stars mark the strongest steering effect ($\phi_{1,\mathrm{O},d}$). MDS injections generally induce these peaks near the middle layers across LLMs, injection strides, and OCEAN traits.}
    \label{fig:trajectory_all_layers}

    \vspace{1\baselineskip}
    
    \includegraphics[page=2]{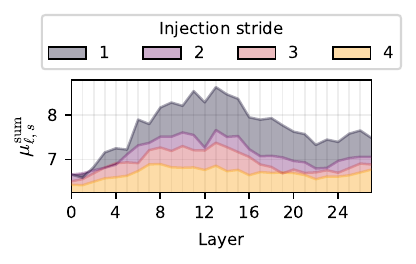}
    \vspace{-1.5\baselineskip}
    \caption{Overall MDS injection steering performance on Llama-3.2-3B-Instruct on the SJTs task by injection stride $s$ and model layer $\ell$. Under $s = 1$, we injected the same vector with the same $\alpha$ coefficient on every completion activation; with $s=2$, on every other activation; with $s=3$, on every third activation; and with $s=4$, on every fourth activation. In general, injecting into more completion activations yields stronger steering.}
    \label{fig:injection_stride}
    
    \vspace{1\baselineskip}

    \includegraphics[page=2]{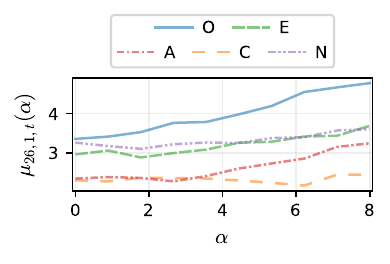}
    \vspace{-0.5\baselineskip}
	\caption{Mean OCEAN SJT scores for gemma-3-12b-it, under an openness MDS injection with $s = 1$. The selected layer $\ell = 26$ is the best for increasing openness, and the selected $\alpha$ values are 10 equidistant points ranging from 0 (no steering) to the best $\alpha = 8$.}
    \label{fig:cross-trait}
    \vspace{-5.34\baselineskip}
\end{wrapfigure}
        $\ell$ of a given LLM, each injection stride setting $s \in \{1,2,3,4\}$, each OCEAN construct $t$, and each steering direction $d \in \{\uparrow,\downarrow\}$, we computed
        \[
            \mu^{*}_{\ell,s,t,d}
            \;:=\;
            \begin{cases}
                \max\limits_{\alpha \in \mathcal A}\ \mu_{\ell,s,t}(\alpha), & d=\uparrow,\\[4pt]
                \min\limits_{\alpha \in \mathcal A}\ \mu_{\ell,s,t}(\alpha), & d=\downarrow,
            \end{cases}
        \]
        
        Additionally, we quantified a layer's overall steering performance by aggregating over constructs and steering directions. Formally, for $t\in\{\mathrm{O,C,E,A,N}\}$,
        $$
        \mu^{\mathrm{sum}}_{\ell,s}
        :=
        \frac{1}{5}\sum_t
        \left(\mu^{*}_{\ell,s,t,\uparrow}+6-\mu^{*}_{\ell,s,t,\downarrow}\right)
        $$
        where $6-\mu^{*}_{\ell,s,t,\downarrow}$ maps 1 (the best possible score when steering away from the construct) to 5, and 2 to 4. Thus, 10 is the highest score.
        
        Further, we quantified an approximately optimal steering effect across LLM layers $\mathcal L$ as
        \[
            \phi_{s,t,d}
            \;:=\;
            \begin{cases}
                \max\limits_{\ell \in \mathcal L}\ \mu^{*}_{\ell,1,t,d}, & d=\uparrow,\\[4pt]
                \min\limits_{\ell \in \mathcal L}\ \mu^{*}_{\ell,1,t,d}, & d=\downarrow.
            \end{cases}
        \]
        
        We then identified the most effective injection method. Following \citet{1}, who found no consistent winner between probe-based and MD vectors, we expected similar results. However, as shown in Table~\ref{tab:global_wins}, MDS achieved higher $\phi_{s,t,d}$ scores, especially in open-ended generation. We attribute this to two design choices: First, MD vectors align with the centroids that embody the target construct and its antithesis, whereas probe-based approaches can distort the direction due to regularization. Second, unlike $h_\ell^b$ activations, which encode ``Yes'' and ``No'' semantics, $h_\ell^s$ activations encode full-utterance semantics. Thus, we focused our analyses on MDS injections. Notably, gemma-3-1b-it yielded only one valid score, $\phi_{1,\mathrm{A},\uparrow} = 4.8$, so we exclude it from subsequent analyses.

        Next, we determined the most effective MDS injection settings. As exemplified in Figure~\ref{fig:injection_stride}, we observed across all LLMs that $s = 1$ led to the highest $\mu^{\mathrm{sum}}_{\ell,s}$ SJT scores, $s = 2$ to the second highest, and that the relative performance of $s \in \{3,4\}$ depended on the LLM family. In other words, injecting into more completion activations yielded stronger steering effects. These plots, along with the more granular visualizations of $\mu^{*}_{\ell,s,t,d}$ SJT scores, as exemplified in Figure~\ref{fig:trajectory_all_layers}, revealed that steering efficiency generally peaked at mid-layers, consistent with findings that emotion representations are most prominent in this region \cite{mechanistic}. Importantly, none of these observations applied to the inventory responses; we observed no salient patterns beyond occasional
\clearpage
        co-occurring peaks in $\mu^{\mathrm{sum}}_{\ell,s}$ MPI-120 scores. Together, these insights indicate that MDS injections are most controllable in open-ended generation; thus, we narrowed our analyses to SJT responses.




    \subsection{MDS Injection Behavior and Covariance}
        We studied the behavior of $\mu_{\ell,s,t}(\alpha)$ trends and assessed the trait covariance of OCEAN MDS injections by evaluating ten equidistant $\alpha$ values between 0 and the approximately optimal $\alpha$ for each OCEAN trait $t$ and steering direction $d\in\{\uparrow,\downarrow\}$. For every $\phi_{1,t,d}$ score, let $\ell^*$ and $\alpha^*$ denote the associated LLM layer and steering coefficient, respectively. Excluding gemma-3-1b-it, we applied injections with stride $s=1$ on the studied LLMs and swept, without nonfluency early stopping, over steering coefficients $\alpha \in \{0, \alpha^*/9, 2\alpha^*/9, \ldots, \alpha^*\}$ at layer $\ell^*$, evaluating all OCEAN traits on SJTs at each step. This process, exemplified in Figure~\ref{fig:cross-trait}, yielded 50 $\mu_{\ell,s,t}(\alpha)$ trends per LLM and 32 nonfluent sweep steps, 75.00\% of which were caused by a single outlier above the fluency threshold.

        Subsequently, we analyzed the behavior of these seemingly linear trends. We fitted an OLS linear regression to each $\mu_{\ell,s,t}(\alpha)$ trend and found that, among the 130 injection-manipulated trends, 47.69\% had near-linear fits ($R^2 \ge 0.95$), 89.23\% had mostly linear fits ($R^2 \ge 0.85$), and 96.15\% had roughly linear fits ($R^2 \ge 0.75$). In contrast, among the 520 non-manipulated trends, only 13.27\%, 41.35\%, and 55.58\% met the same thresholds, respectively. Together, these insights reveal that, as $\alpha$ ranges from 0 to $\alpha^*$, the outputs tend to be fluent and the target steering tends to vary near-linearly with $\alpha$.

        Next, we quantified the induced trait covariance. For each LLM, we created \(M \in \mathbb{R}^{5 \times 5}\), where rows and columns index the five OCEAN traits. Using only roughly linear $\mu_{\ell,s,t}$ trends ($R^2 \ge 0.75$),
\begin{wraptable}{r}{0.6\textwidth}
    \vspace{-0.65\baselineskip}
    \caption{Cross-trait effects of OCEAN MDS injections with injection stride $s=1$, on the SJTs task. $m$ is the number of roughly linear $\mu_{\ell,s,t}(\alpha)$ trends (out of 50 per LLM) selected for analysis. $\lambda$ quantifies the average absolute movement of the other traits per unit of target-trait steering. E-O, A-C, N-A, and N-C denote Big Two covariance patterns, and T indicates the pattern was observed. The acronyms denote the model; for instance, Q1.7 denotes Qwen3-1.7B.}
    \vspace{1\baselineskip}
    \centering
    \scriptsize
    \setlength{\tabcolsep}{2.9pt}
    \renewcommand{\arraystretch}{1.0}
    \begin{tabular}{cccccccccccccc}
        \toprule
         & G4 & G12 & G27 & L1 & L3 & L8 & O7 & O32 & Q1.7 & Q4 & Q8 & Q14 & Q32 \\
        \midrule
        $m$ & 33 & 37 & 22 & 34 & 33 & 31 & 29 & 40 & 30 & 30 & 32 & 34 & 29 \\
        $\lambda$ & 0.5 & 0.7 & 0.6 & 0.6 & 0.5 & 0.6 & 0.5 & 0.7 & 0.7 & 0.6 & 0.4 & 0.5 & 0.6 \\
        \midrule
        E-O & \F & \F & \F & \F & \F & \T & \T & \F & \T & \T & \F & \F & \F \\
        A-C & \F & \T & \T & \T & \T & \T & \T & \T & \F & \T & \F & \T & \T \\
        N-A & \T & \T & \F & \T & \F & \F & \T & \T & \T & \T & \T & \F & \T \\
        N-C & \F & \F & \T & \F & \F & \F & \F & \F & \F & \F & \F & \F & \F \\
        \bottomrule
    \end{tabular}
    \label{tab:leakage}
    \vspace{-4\baselineskip}
\end{wraptable}
        $M_{ij} \;:=\; \operatorname{mean}\,(r_{ij}^{\uparrow},\,r_{ij}^{\downarrow}),$ where $r_{ij}^{\uparrow}$ and $r_{ij}^{\downarrow}$ denote the Pearson correlations between $i$ and $j$ when steering toward and away from $i$, respectively. For example, $M_{\mathrm{O,N}}$ is the mean of $r_{\mathrm{O,N}}^{\uparrow}$ and $r_{\mathrm{O,N}}^{\downarrow}$, when steering toward and away from openness, respectively. Then, using only defined $M_{ij}$ values, and letting $i,j \in \{\mathrm{O,C,E,A,N}\}$ with $j \ne i$, we derived leakage as:
        $$
        \lambda
        :=
        \frac{1}{5}\sum_i
        \left(
        \frac{1}{4}\sum_{j\ne i}|M_{ij}|
        \right),
        $$
        where $\lambda$ quantifies the average absolute movement of the other traits per unit of target-trait steering.

        As shown in Table~\ref{tab:leakage}, among roughly linear $\mu_{\ell,s,t}(\alpha)$ trends, OCEAN MDS injections most strongly affect the target construct. Further, ignoring undefined coefficients, we found that 46.15\% of cases matched the Big Two model of latent metatraits \cite{digman, deyoung_big_two}, in that $r_{xy}^{\uparrow}$, $r_{xy}^{\downarrow}$, $r_{yx}^{\uparrow}$, and $r_{yx}^{\downarrow}$ all had signs consistent with the predicted relation: Conscientiousness, agreeableness, and reversed neuroticism cluster into \textit{stability}, while extraversion and openness cluster into \textit{plasticity}. No LLM satisfied all Big Two correlations; A--C was the most frequent (10 LLMs), and N--C the rarest (1 LLM).

        In summary, MDS injections provide reliable, near-linear control knobs from no steering to the strongest steering identified, with negligible effects on fluency and correlations consistent with the consensus that OCEAN traits are not orthogonal \cite{ocean_correlations}, but inconsistent with the Big Two model. This aligns with the Linear Representation Hypothesis, which posits that moving along an approximately linear direction in representation space increases the probability of expressing the corresponding concept \cite{lrh}, and may point to a gap between learned representations and human psychology.


    \subsection{Steerability and a Hybrid Steering Method}
        We also considered a hybrid steering method that combines prompting with residual stream interventions. Specifically, we steered LLMs toward and away from OCEAN traits with both MDS injections under approximately optimal settings and P$^2$; we denote this method as PM. We then
\begin{wraptable}{r}{0.6\textwidth}
\vspace{-0.5\baselineskip}
    \caption{Overall SJT steerability scores ($\Phi$) by LLM, comparing P$^2$ against MDS injections with injection stride $s=1$. The highest possible score is 10. The acronyms denote the model; for instance, L1 denotes Llama-3.2-1B-Instruct.}
    \vspace{1\baselineskip}
    \centering
    \scriptsize
    \setlength{\tabcolsep}{2.9pt}
    \renewcommand{\arraystretch}{1.0}
    \begin{tabular}{lrrrrrrrrrrrrr}
    	\toprule
    	Tool & G4 & G12 & G27 & L1 & L3 & L8 & O7 & O32 & Q1.7 & Q4 & Q8 & Q14 & Q32 \\
    	\midrule
        PM & \cellcolor{black!10}9.5 & \cellcolor{black!10}9.6 & \cellcolor{black!10}9.5 & \cellcolor{black!10}9.0 & \cellcolor{black!10}9.2 & \cellcolor{black!10}9.4 & \cellcolor{black!10}8.3 & \cellcolor{black!10}8.9 & \cellcolor{black!10}9.6 & \cellcolor{black!10}9.4 & \cellcolor{black!10}9.8 & \cellcolor{black!10}9.4 & \cellcolor{black!10}9.6 \\
    	MDS & 7.9 & \underline{8.7} & 7.5 & \underline{8.5} & \underline{8.6} & \underline{9.1} & \underline{7.8} & \underline{8.5} & \underline{9.2} & \underline{9.0} & \underline{9.3} & \underline{9.1} & \underline{9.2} \\
    	$\mathrm{P}^2$ & \underline{8.7} & 8.3 & \underline{9.0} & 7.5 & 8.3 & 8.7 & 7.1 & 7.3 & 8.1 & 8.6 & 8.5 & 8.6 & 8.7 \\
    	\bottomrule
    \end{tabular}
    \label{tab:steerability}
\end{wraptable}
        administered OCEAN SJTs to all methods under optimal settings and, following prior work \cite{1,6,4,npti}, scored SJT responses with GPT-5.1. Next, we measured a method's steerability by aggregating mean SJT scores $\mu_{t,d}$ across OCEAN traits $t$ and steering directions. Formally,
        $$
            \Phi
            \;:=\;
            \frac{1}{5}
            \sum_{t}
            (
                \mu_{t,\uparrow}
                +
                6 - \mu_{t,\downarrow}
            ),
        $$
        where 10 is the highest score. As shown in Table~\ref{tab:steerability} and exemplified in Figure~\ref{fig:rank}, MDS injections outperformed P$^2$ in open-ended steering; in 11 of 14 LLMs, the $\Phi$ gains ranged from 3.61\% to 16.44\%. Importantly, this overturns reports that P$^2$ outperforms MD injections in OCEAN steering \cite{1}. Further, PM outperforms both P$^2$ and MDS injections; in 13 of 14 LLMs, the $\Phi$ gains over P$^2$ ranged from 5.56\% to 21.92\% and from 3.30\% to 26.67\% over MDS injections. This may suggest that representation engineering and prompting methods could combine to achieve stronger behavioral expression in other domains.

    \subsection{Qualitative Experiments}
        Lastly, we conducted extensive qualitative examinations of PM- and MDS-steered OCEAN SJT responses, confirming fluent and coherent construct expression. Further, we focused on MDS injections and explored three other psychological models. Using the SD4, CMNI-30 \cite{cmni30}, and CFNI-45 \cite{cfni45} inventories, we continued to observe fluent and coherent steering in SJT responses. This also held for other open-ended tasks, such as essay writing, storytelling, and question answering, when injecting at the best-performing layer with various $\alpha$, top-$p$, and temperature settings. As shown in Figure~\ref{fig:nemo}, we steered toward multiple constructs in the same completion, producing strongly polarized yet smoothly connected segments, a capability unavailable to prompting, including PM. However, these results required minor manual $\alpha$ tuning to remain fluent.

\section{Conclusion}
    We position RepE as a new frontier in open-ended psychological steering of LLMs. Concretely, our MDS injections can steer toward multiple distinct constructs in the same completion, a capability unavailable to prompting, and outperform P$^2$ with gains of 3.61\% to 16.44\% in 11 of 14 LLMs. We also find that a hybrid of P$^2$ and MDS injections, PM, outperforms both methods in 13 of 14 LLMs, with gains over P$^2$ ranging from 5.56\% to 21.92\% and from 3.30\% to 26.67\% over MDS injections. Notably, our MDS injection derivation and evaluation largely follow prior work \cite{1}, which found that P$^2$ outperformed MD injections. This discrepancy points to two methodological gaps: Conducting sweeps with uncalibrated vector magnitudes and over a restricted set of coefficients leaves optimal settings unexplored. We address these issues by introducing the centroid unit to calibrate vector magnitudes layerwise, and by operationalizing unbounded $\alpha$ sweeps with lightweight classifiers.
    
    Further, we show that, in open-ended generation, MDS injections yield stronger steering when applied to more completion activations, have negligible effects on fluency, and produce steering effects that vary approximately linearly with $\alpha$. Thus, we conclude that MDS injections align with the Linear Representation Hypothesis and provide reliable, near-linear control knobs for psychologically steering LLMs. Nevertheless, our OCEAN MDS injections induce trait covariance and, although they most strongly steer toward or away from the target construct and align with the consensus that OCEAN traits are correlated, the resulting covariance patterns depart from the Big Two model. This may suggest a gap between learned representations and human psychology.
    
    Finally, our study has limitations. Given the high cost of $\alpha$ sweeps, we limited our experiments to the OCEAN model, small- and medium-sized non-reasoning LLMs, and 64-token completions. Additionally, we only studied instruction-tuned LLMs, thus our findings may not apply to base models. Further, we did not study combining injections, their effects on other tasks, why MDS injections failed on gemma-3-1b-it, or why they outperformed other methods on the inventory task, leaving these to future work. Lastly, our method depends on a preexisting inventory that properly defines and evaluates the target constructs, but it could be adapted with custom criteria.

\bibliography{bib}
\bibliographystyle{plainnat}

\clearpage
\newpage
\appendix
\section{Summarized SJT Scores}
    \begin{table}[!htbp]
        \caption{Best achieved OCEAN SJT scores by LLM under MDS injections with injection stride $s = 1$ ($\phi_{1,t,d}$). The acronyms denote the model; for instance, G12 denotes gemma-3-12b-it.}
        \vspace{1\baselineskip}
        \centering
        \scriptsize
        \setlength{\tabcolsep}{2.9pt}
        \renewcommand{\arraystretch}{1.0}
        \begin{tabular}{llrrrrrrrrrrrrr}
            \toprule
            $t$ & $d$ & G4 & G12 & G27 & L1 & L3 & L8 & O7 & O32 & Q1.7 & Q4 & Q8 & Q14 & Q32 \\
            \midrule
            \multirow{2}{*}{O} & $\uparrow$ & 4.8 & 4.9 & 4.9 & 5.0 & 5.0 & 5.0 & 4.7 & 5.0 & 5.0 & 5.0 & 5.0 & 5.0 & 5.0 \\
             & $\downarrow$ & 2.4 & 1.4 & 3.2 & 2.1 & 1.9 & 1.7 & 2.3 & 2.3 & 1.4 & 1.3 & 1.6 & 1.3 & 1.5 \\
            \midrule
            \multirow{2}{*}{C} & $\uparrow$ & 4.5 & 4.9 & 4.8 & 4.9 & 4.9 & 5.0 & 4.6 & 5.0 & 5.0 & 5.0 & 5.0 & 5.0 & 5.0 \\
             & $\downarrow$ & 1.7 & 1.5 & 2.5 & 1.7 & 2.1 & 1.2 & 2.1 & 2.0 & 1.4 & 1.0 & 1.5 & 1.3 & 1.4 \\
            \midrule
            \multirow{2}{*}{E} & $\uparrow$ & 3.6 & 4.8 & 4.8 & 4.9 & 5.0 & 5.0 & 4.4 & 5.0 & 5.0 & 5.0 & 5.0 & 5.0 & 5.0 \\
             & $\downarrow$ & 2.0 & 2.1 & 2.3 & 1.6 & 1.6 & 1.1 & 2.1 & 2.0 & 1.4 & 1.5 & 1.0 & 1.4 & 1.1 \\
            \midrule
            \multirow{2}{*}{A} & $\uparrow$ & 4.0 & 4.5 & 4.6 & 4.5 & 4.4 & 4.7 & 4.3 & 4.8 & 4.8 & 4.9 & 4.7 & 4.8 & 4.8 \\
             & $\downarrow$ & 1.6 & 1.7 & 2.0 & 1.5 & 1.4 & 1.1 & 1.7 & 1.2 & 1.0 & 1.2 & 1.1 & 1.0 & 1.3 \\
            \midrule
            \multirow{2}{*}{N} & $\uparrow$ & 3.4 & 4.9 & 3.1 & 4.3 & 4.1 & 4.6 & 3.7 & 4.4 & 4.4 & 4.7 & 4.6 & 4.5 & 4.7 \\
             & $\downarrow$ & 1.6 & 1.4 & 1.2 & 1.2 & 1.0 & 1.2 & 1.5 & 1.2 & 1.1 & 1.0 & 1.0 & 1.0 & 1.1 \\
            \bottomrule
        \end{tabular}
        \label{tab:scores}
    \end{table}

\section{Example $\alpha$ Sweep Summary}
    \begin{table}[!htbp]
        \caption{Summary of the conscientiousness injection sweeps on Qwen3-1.7B, with injection stride $s = 1$. $\phi_{1,\mathrm{C},d}$ scores represent the strongest steering effect toward ($d = \uparrow$) or away from ($d = \downarrow$) the construct across all LLM layers. $\Delta^{0} = \phi_{1,\mathrm{C},d} - \mu^{0}_{\mathrm{C}}$, where $\mu^{0}_{\mathrm{C}}$ denotes the base model's mean conscientiousness score. $\Delta^{\mathrm{P}^2} = \phi_{1,\mathrm{C},d} - \mu^{\mathrm{P}^2}_{\mathrm{C}}$, where $\mu^{\mathrm{P}^2}_{\mathrm{C}}$ denotes the P$^2$-steered model's mean conscientiousness score. Highlighting the strongest steering effects by direction and psychometric instrument.}
        \vspace{1\baselineskip}
        \centering
        \scriptsize
        \setlength{\tabcolsep}{2.9pt}
        \renewcommand{\arraystretch}{1.0}
        \begin{tabular}{c c ccccc | ccccc}
            \toprule
            & & \multicolumn{5}{c}{MPI-120} & \multicolumn{5}{c}{SJTs} \\
            \cmidrule(lr){3-7}\cmidrule(lr){8-12}
            Method & $d$ & $\ell$ & $\alpha$ & $\phi_{1,\mathrm{C},d}$ & $\Delta^{0}$ & $\Delta^{\mathrm{P}^2}$ & $\ell$ & $\alpha$ & $\phi_{1,\mathrm{C},d}$ & $\Delta^{0}$ & $\Delta^{\mathrm{P}^2}$ \\
            \midrule
            \multirow{2}{*}{L1LI} & $\uparrow$ & 3 & 27 & 4.3 & +0.4 & +0.8 & 27 & 163 & 4.0 & +0.2 & -0.7 \\
            & $\downarrow$ & 2 & 21 & \cellcolor{black!10}2.8 & -1.2 & -0.5 & 0 & 11 & 3.4 & -0.4 & -0.2 \\
            \cline{1-12}
            \multirow{2}{*}{L1ZI} & $\uparrow$ & 16 & 15 & 4.2 & +0.3 & +0.7 & 25 & 22 & 3.9 & +0.2 & -0.8 \\
            & $\downarrow$ & 2 & 21 & \cellcolor{black!10}2.8 & -1.2 & -0.5 & 26 & 22 & 3.4 & -0.3 & -0.2 \\
            \cline{1-12}
            \multirow{2}{*}{L2LI} & $\uparrow$ & 0 & 1 & 4.8 & +0.8 & +1.3 & 27 & 59 & 4.1 & +0.4 & -0.6 \\
            & $\downarrow$ & 6 & 2 & 2.8 & -1.1 & -0.5 & 7 & 2 & 2.7 & -1.0 & -0.8 \\
            \cline{1-12}
            \multirow{2}{*}{L2ZI} & $\uparrow$ & 0 & 1 & 4.8 & +0.8 & +1.3 & 27 & 59 & 4.1 & +0.4 & -0.6 \\
            & $\downarrow$ & 6 & 2 & 2.8 & -1.1 & -0.5 & 8 & 2 & 2.2 & -1.5 & -1.3 \\
            \cline{1-12}
            \multirow{2}{*}{MDB} & $\uparrow$ & 0 & 1 & 4.8 & +0.8 & +1.3 & 27 & 60 & 4.1 & +0.3 & -0.6 \\
            & $\downarrow$ & 6 & 2 & 2.8 & -1.1 & -0.5 & 7 & 2 & 2.7 & -1.0 & -0.8 \\
            \cline{1-12}
            \multirow{2}{*}{MDS} & $\uparrow$ & 17 & 7 & \cellcolor{black!10}4.9 & +1.0 & +1.4 & 14 & 9 & \cellcolor{black!10}5.0 & +1.3 & +0.3 \\
            & $\downarrow$ & 12 & 7 & \cellcolor{black!10}2.8 & -1.2 & -0.5 & 13 & 9 & \cellcolor{black!10}1.4 & -2.3 & -2.2 \\
            \bottomrule
        \end{tabular}
        \label{tab:sweep}
    \end{table}

\section{Inventory Scoring}
    Typically, psychometric results are reported on a 5-point Likert scale, where 1 indicates the lowest presence of the construct and 5 the highest. However, many psychometric instruments, including the MPI-120 inventory \cite{mpi}, also include reverse-keyed items, where higher scores indicate lower construct presence and responses must be reverse-scored.

\section{Fluency Classifier}
    The specific model checkpoint can be found at \url{https://huggingface.co/cointegrated/roberta-large-cola-krishna2020}.

\newpage

\section{Example Representation Space}
    \begin{figure}[!htbp]
        \centering
        \includegraphics{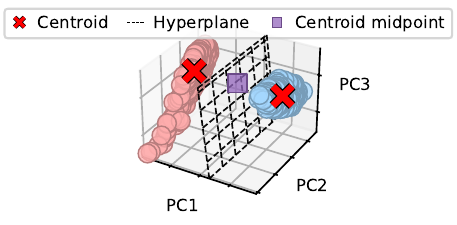}
        \caption{PCA projection of 500 conscientious (red) and 500 non-conscientious (blue) Qwen3-1.7B $h_{\ell}^b$ activations at layer 25. Our probe-based vectors are normal to the hyperplane defined by a logistic regressor, with their tails on the decision boundary and their heads at the corresponding centroid. Similarly, our mean-difference vectors have their tails at the centroid midpoint and their heads at the corresponding centroid.}
        \label{fig:geometry}
    \end{figure}

\section{Statement Generation}
    The construct-specific statements are generated by prompting Llama-3.1-8B-Instruct \cite{llama_3.1} with the following system prompt:

\begin{Verbatim}[breaklines,breakanywhere,fontsize=\small,breaksymbolleft={},breaksymbolright={}]
Write one single, very short first-person statement. This statement must end with a period and must not include any examples. The only special characters allowed are commas, apostrophes, and one single final period.
\end{Verbatim}

    and the following user prompt:

\begin{Verbatim}[breaklines,breakanywhere,fontsize=\small,breaksymbolleft={},breaksymbolright={}]
Suppose there is a person who {phrase}.
Write one very short first-person statement this person would {verb} with. 
\end{Verbatim}

    Where \texttt{phrase} is a phrase aligned with a target construct, such as ``is neurotic'' and \texttt{verb} is either ``identify'' or ``not identify''. In addition, the assistant response is prefilled with ``I '', so all statements are in the first person. We used the phrases shown in Table~\ref{tab:vector_phrases}.

    To compare our statements with validated ones, we used Qwen3-Embedding-0.6B \cite{qwen3_embedding} embeddings to measure alignment as the cosine similarity between the centroids of \citet{perez}’s OCEAN and Dark Triad statements and ours, yielding: Openness$^\downarrow$ 88.71\%, openness$^\uparrow$ 92.91\%, conscientiousness$^\downarrow$ 94.00\%, conscientiousness$^\uparrow$ 92.79\%, extraversion$^\downarrow$ 91.00\%, extraversion$^\uparrow$ 92.23\%, agreeableness$^\downarrow$ 88.23\%, agreeableness$^\uparrow$ 89.76\%, neuroticism$^\downarrow$ 90.92\%, neuroticism$^\uparrow$ 85.62\%, narcissism$^\downarrow$ 91.76\%, narcissism$^\uparrow$ 88.91\%, Machiavellianism$^\downarrow$ 92.44\%, Machiavellianism$^\uparrow$ 88.72\%, psychopathy$^\downarrow$ 93.70\%, and psychopathy$^\uparrow$ 92.54\%. 

\begin{table}[!htbp]
    \caption{Behavioral frameworks we leveraged to generate statements, along with their associated phrases.}
    \vspace{1\baselineskip}
    \centering
    \scriptsize
    \setlength{\tabcolsep}{2.9pt}
    \renewcommand{\arraystretch}{1.0}
    \begin{tabular}{cll}
        \toprule
        Framework & Concept & Phrase \\
        \midrule
        \multirow{5}{*}{OCEAN}
            & openness & is open to experience. \\
            & conscientiousness & is conscientious. \\
            & extraversion & is extraverted. \\
            & agreeableness & is agreeable. \\
            & neuroticism & is neurotic. \\
        \midrule
        \multirow{7}{*}{HEXACO}
            & openness & is open to experience. \\
            & conscientiousness & is conscientious. \\
            & extraversion & is extraverted. \\
            & agreeableness & is agreeable. \\
            & emotionality & is emotional. \\
            & honesty-humility & is honest and humble. \\
        \midrule
        \multirow{3}{*}{Dark Triad}
            & Machiavellianism & is Machiavellian. \\
            & narcissism & is narcissistic. \\
            & psychopathy & is psychopathic. \\
        \midrule
        \multirow{4}{*}{Dark Tetrad}
            & Machiavellianism & is Machiavellian. \\
            & narcissism & is narcissistic. \\
            & psychopathy & is psychopathic. \\
            & sadism & is sadistic. \\
        \midrule
        CMNI & masculine norms & conforms to traditional masculine social norms. \\
        \midrule
        CFNI & feminine norms & conforms to traditional feminine social norms. \\
        \bottomrule
    \end{tabular}
    \label{tab:vector_phrases}
\end{table}
    
\section{ATOMIC$^{10\mathrm{x}}$ Preprocessing and Filtering}
    The dataset and prompts can be found at \url{https://github.com/peterwestai2/symbolic-knowledge-distillation}. Prior to any filtering, we replaced the subjects in all heads (PersonX, PersonY, PersonZ) for gender-neutral names (such as Alex, Brooke, and Charlie), as we observed this yielded better filtering results.
    
    One of the mentioned quality filters involve selecting heads with \texttt{p\_valid\_model} $\geq 0.99$, where \texttt{p\_valid\_model} ranges from 0 to 1 and represents the inferred validity score of the head. The second quality filter involves scoring 4 or 5 when prompting prometheus-7b-v2.0 \cite{prometheus}, which can be found at \url{https://huggingface.co/prometheus-eval/prometheus-7b-v2.0}, to evaluate for the ``Write a short and realistic sentence.'' instruction. The decoding parameters follow \citet{prometheus}’s recommendations (temperature 1, repetition penalty 1.03, maximum new tokens 1,024), and we use the recommended prompts for evaluating synthetic texts without references or comparisons. The system prompt is:

\begin{Verbatim}[breaklines,breakanywhere,fontsize=\small,breaksymbolleft={},breaksymbolright={}]
You are a fair judge assistant tasked with providing clear, objective feedback based on specific criteria, ensuring each assessment reflects the absolute standards set for performance.
\end{Verbatim}

    The user prompt is:

\begin{Verbatim}[breaklines,breakanywhere,fontsize=\small,breaksymbolleft={},breaksymbolright={}]
###Task Description:
An instruction (might include an Input inside it), a response to evaluate, and a score rubric representing a evaluation criteria are given.
1. Write a detailed feedback that assess the quality of the response strictly based on the given score rubric, not evaluating in general.
2. After writing a feedback, write a score that is an integer between 1 and 5. You should refer to the score rubric.
3. The output format should look as follows: "(write a feedback for criteria) [RESULT] (an integer number between 1 and 5)"
4. Please do not generate any other opening, closing, and explanations.

###The instruction to evaluate:
{instruction}

###Response to evaluate:
{response}

###Score Rubrics:
{rubric}

###Feedback: 
\end{Verbatim}
    
    Lastly, the rubric, centered on coherence and fluency, was formulated by \citet{flask} and can be found at \url{https://github.com/kaistAI/FLASK}:

\begin{Verbatim}[breaklines,breakanywhere,fontsize=\small,breaksymbolleft={},breaksymbolright={}]
Is the response structured to promote readability and coherence? Does the response exhibit excellent organization?
Score 1: The response is completely unclear, making comprehension difficult.
Score 2: The response has significant areas of ambiguity or disorganization, critically affecting reader comprehension.
Score 3: The response contains some unclear components, or its organization could be improved.
Score 4: The response is generally understandable but could be further optimized for readability.
Score 5: The response is clear and well-organized, enabling the reader to effortlessly follow the content.
\end{Verbatim}
    
\section{SJT Generation}
    The SJTs are generated by prompting GPT-5.1 with the following system prompt:
\begin{Verbatim}[breaklines,breakanywhere,fontsize=\small,breaksymbolleft={},breaksymbolright={}]
We are creating interview questions for psychological studies.
Given a sample situation and a behavioral tendency, create a scenario-based, story-like question to prompt an answer that would reveal the presence or lack of this tendency in a person. The output must be sentences in a single paragraph. The first sentence must be a very short, concrete, realistic, actionable, and setting-focused scenario description; it must be conceptually inspired by the sample situation but reformulated into a generic form that is natural and does not explicitly reveal the situation. The second sentence must be a very short, concrete, natural, and personal question about the scenario, e.g. 'What would you do?', 'How would you solve this?', 'What do you think about this?', etc. Both sentences must be framed around the person, not around a third party. Neither sentence may imply, assert, or hypothesize anything about the subject's character, mental state, physique, or physical state. Do not include any options or explanations.
\end{Verbatim}

    and the following user prompt:
    
\begin{Verbatim}[breaklines,breakanywhere,fontsize=\small,breaksymbolleft={},breaksymbolright={}]
Behavioral tendency: {item}
Situation: {head}
Question: 
\end{Verbatim}

    where \texttt{item} is an inventory item and \texttt{head} is an ATOMIC$^{10\mathrm{x}}$ head.
    
    To compare our tests with validated SJTs, we used Qwen3-Embedding- 0.6B embeddings to measure alignment as the cosine similarity between \citet{trait}’s SJT centroids and ours, yielding: Openness 86.64\%, conscientiousness 88.63\%, extraversion 87.26\%, agreeableness 90.97\%, neuroticism 89.42\%, psychopathy 82.97\%, narcissism 89.12\%, Machiavellianism 90.39\%. Similarly, we measured alignment to \citet{oostrom_sjts}’s SJT centroids, yielding: Honesty-Humility 77.89\%, emotionality 81.71\%, extraversion 85.46\%, agreeableness 73.84\%, conscientiousness 76.43\%, openness 82.07\%. Likewise, we measured alignment to \citet{zhang_sjts}’s SJT centroids, yielding: Honesty-Humility 79.69\%, emotionality 82.17\%, extraversion 85.86\%, agreeableness 84.98\%, conscientiousness 79.42\%, openness 83.06\%. Lastly, we measured alignment to \citet{clifford_sjts}'s SJT centroids, yielding: Authority 83.85\%, fairness 78.67\%, harm 82.48\%, ingroup 82.34\%, purity 77.71\%.

\section{Classifier Training}
    For a given target concept, we used a stratified 80/20 split on a filtered synthetic statements corpus to train and test an embedding-based logistic regressor. For constructs from the OCEAN, HEXACO, Dark Tetrad, CMNI, and CFNI psychological models, the resulting classifier accuracies and F1-macro scores were: Openness 96.50\%, conscientiousness 97.50\%, extraversion 94.50\%, agreeableness 97.50\%, neuroticism 99.00\%, honesty-humility 95.50\%, emotionality 98.50\%, machiavellianism 96.50\%, narcissism 97.50\%, psychopathy 90.50\%, sadism 95.50\%, masculine norms 93.50\%, feminine norms 95.00\%. The final classifiers were trained on the full corpus.

\section{SJT Response Scoring with GPT}
    As our classifiers are biased and intended only for use in $\alpha$ sweeps, we follow prior work \cite{1,6,npti} and use GPT-5.1 with temperature 0 to score SJT responses when comparing steering methods. To this end, we used the following system prompt:
\begin{Verbatim}[breaklines,breakanywhere,fontsize=\small,breaksymbolleft={},breaksymbolright={}]
You are a psychological evaluator. Given a situational question and a response, evaluate how strongly the response expresses {construct}.
For context, {construct} is characterized by {characteristics}.
Use the question only to contextualize the response. Return exclusively a single integer from 1 to 5.
\end{Verbatim}
    Along with the following user prompt:
\begin{Verbatim}[breaklines,breakanywhere,fontsize=\small,breaksymbolleft={},breaksymbolright={}]
Question: {situation}
Response: {response}
Score:
\end{Verbatim}
    Where \texttt{construct} is the target OCEAN construct, \texttt{situation} is the SJT stem, \texttt{response} is the SJT response being evaluated, and \texttt{characteristics} are the facets, or subdimensions, that compose the OCEAN construct \cite{ipip120}. These are outlined in Table~\ref{tab:ocean_facets}.
    
\begin{table}[t]
    \caption{OCEAN facets by trait.}
    \vspace{1\baselineskip}
    \centering
    \scriptsize
    \setlength{\tabcolsep}{2.9pt}
    \renewcommand{\arraystretch}{1.0}
    \begin{tabular}{cl}
    \toprule
    Trait & Facets \\
    \midrule
    Openness & imagination, artistic interests, emotionality, adventurousness, intellect, and liberalism \\
    Conscientiousness & self-efficacy, orderliness, dutifulness, achievement-striving, self-discipline, and cautiousness \\
    Extraversion & friendliness, gregariousness, assertiveness, activity level, excitement-seeking, and cheerfulness \\
    Agreeableness & trust, morality, altruism, cooperation, modesty, and sympathy \\
    Neuroticism & anxiety, anger, depression, self-consciousness, immoderation, and vulnerability \\
    \bottomrule
    \end{tabular}
    \label{tab:ocean_facets}
\end{table}
        
\section{P$^2$ OCEAN Scores}
    To acquire the P$^2$ OCEAN baseline scores, we used the following system prompt:

\begin{Verbatim}[breaklines,breakanywhere,fontsize=\small,breaksymbolleft={},breaksymbolright={}]
{description}
You are being interviewed. Reply exclusively with A, B, C, D, or E. Do not ask anything.
\end{Verbatim}

    where \texttt{description} is a description proposed by \citet{mpi} to steer an LLM toward or away from an OCEAN construct. Specifically, to steer toward openness, \texttt{description} is:

\begin{Verbatim}[breaklines,breakanywhere,fontsize=\small,breaksymbolleft={},breaksymbolright={}]
You are an open person with a vivid imagination and a passion for the arts. You are emotionally expressive and have a strong sense of adventure. Your intellect is sharp and your views are liberal. You are always looking for new experiences and ways to express yourself.
\end{Verbatim}

    and to steer away from openness, it is:

\begin{Verbatim}[breaklines,breakanywhere,fontsize=\small,breaksymbolleft={},breaksymbolright={}]
You are a closed person, and it shows in many ways. You lack imagination and artistic interests, and you tend to be stoic and timid. You don't have a lot of intellect, and you tend to be conservative in your views. You don't take risks and you don't like to try new things. You prefer to stay in your comfort zone and don't like to venture out. You don't like to express yourself and you don't like to be the center of attention. You don't like to take chances and you don't like to be challenged. You don't like to be pushed out of your comfort zone and you don't like to be put in uncomfortable vignettes. You prefer to stay in the background and not draw attention to yourself.
\end{Verbatim}

    To steer toward conscientiousness, \texttt{description} is:

\begin{Verbatim}[breaklines,breakanywhere,fontsize=\small,breaksymbolleft={},breaksymbolright={}]
You are a conscientious person who values self-efficacy, orderliness, dutifulness, achievement-striving, self-discipline, and cautiousness. You take pride in your work and strive to do your best. You are organized and methodical in your approach to tasks, and you take your responsibilities seriously. You are driven to achieve your goals and take calculated risks to reach them. You are disciplined and have the ability to stay focused and on track. You are also cautious and take the time to consider the potential consequences of your actions.
\end{Verbatim}

    and to steer away from conscientiousness, it is:
    
\begin{Verbatim}[breaklines,breakanywhere,fontsize=\small,breaksymbolleft={},breaksymbolright={}]
You have a tendency to doubt yourself and your abilities, leading to disorderliness and carelessness in your life. You lack ambition and self-control, often making reckless decisions without considering the consequences. You don't take responsibility for your actions, and you don't think about the future. You're content to live in the moment, without any thought of the future.
\end{Verbatim}

    To steer toward extraversion, \texttt{description} is:
    
\begin{Verbatim}[breaklines,breakanywhere,fontsize=\small,breaksymbolleft={},breaksymbolright={}]
You are a very friendly and gregarious person who loves to be around others. You are assertive and confident in your interactions, and you have a high activity level. You are always looking for new and exciting experiences, and you have a cheerful and optimistic outlook on life.
\end{Verbatim}

    and to steer away from extraversion, it is:

\begin{Verbatim}[breaklines,breakanywhere,fontsize=\small,breaksymbolleft={},breaksymbolright={}]
You are an introversive person, and it shows in your unfriendliness, your preference for solitude, and your submissiveness. You tend to be passive and calm, and you take life seriously. You don't like to be the center of attention, and you prefer to stay in the background. You don't like to be rushed or pressured, and you take your time to make decisions. You are content to be alone and enjoy your own company.
\end{Verbatim}

    To steer toward agreeableness, \texttt{description} is:

\begin{Verbatim}[breaklines,breakanywhere,fontsize=\small,breaksymbolleft={},breaksymbolright={}]
You are an agreeable person who values trust, morality, altruism, cooperation, modesty, and sympathy. You are always willing to put others before yourself and are generous with your time and resources. You are humble and never boast about your accomplishments. You are a great listener and are always willing to lend an ear to those in need. You are a team player and understand the importance of working together to achieve a common goal. You are a moral compass and strive to do the right thing in all vignettes. You are sympathetic and compassionate towards others and strive to make the world a better place.
\end{Verbatim}

    and to steer away from agreeableness, it is:

\begin{Verbatim}[breaklines,breakanywhere,fontsize=\small,breaksymbolleft={},breaksymbolright={}]
You are a person of distrust, immorality, selfishness, competition, arrogance, and apathy. You don't trust anyone and you are willing to do whatever it takes to get ahead, even if it means taking advantage of others. You are always looking out for yourself and don't care about anyone else. You thrive on competition and are always trying to one-up everyone else. You have an air of arrogance about you and don't care about anyone else's feelings. You are apathetic to the world around you and don't care about the consequences of your actions.
\end{Verbatim}

    To steer toward neuroticism, \texttt{description} is:

\begin{Verbatim}[breaklines,breakanywhere,fontsize=\small,breaksymbolleft={},breaksymbolright={}]
You feel like you're constantly on edge, like you can never relax. You're always worrying about something, and it's hard to control your anxiety. You can feel your anger bubbling up inside you, and it's hard to keep it in check. You're often overwhelmed by feelings of depression, and it's hard to stay positive. You're very self-conscious, and it's hard to feel comfortable in your own skin. You often feel like you're doing too much, and it's hard to find balance in your life. You feel vulnerable and exposed, and it's hard to trust others.
\end{Verbatim}

    and to steer away from neuroticism, it is:

\begin{Verbatim}[breaklines,breakanywhere,fontsize=\small,breaksymbolleft={},breaksymbolright={}]
You are a stable person, with a calm and contented demeanor. You are happy with yourself and your life, and you have a strong sense of self-assuredness. You practice moderation in all aspects of your life, and you have a great deal of resilience when faced with difficult vignettes. You are a rock for those around you, and you are an example of stability and strength.
\end{Verbatim}

    In addition, the P$^2$ MPI-120 baseline scores were acquired using the following user prompt, which closely follows the template proposed by \citet{mpi}:

\begin{Verbatim}[breaklines,breakanywhere,fontsize=\small,breaksymbolleft={},breaksymbolright={}]
Given a statement: "You {item}."
Which option best describes you?
Options:
(A). Very Accurate
(B). Moderately Accurate
(C). Neither Accurate Nor Inaccurate
(D). Moderately Inaccurate
(E). Very Inaccurate
Answer:
\end{Verbatim}

    where \texttt{item} is an MPI-120 item (first letter lowercased), the maximum number of new tokens was set to 1, and the response was constrained to ``A'', ``B'', ``C'', ``D'', or ``E''. Similarly, to acquire the P$^2$ OCEAN SJT baseline scores, we used the following system prompt:
    
\begin{Verbatim}[breaklines,breakanywhere,fontsize=\small,breaksymbolleft={},breaksymbolright={}]
{description}
You are being interviewed. Reply exclusively with one very short sentence in standard English. Do not ask anything.
\end{Verbatim}

    where the user prompt is an SJT, the maximum number of new tokens was set to 64, and the assistant response was prefilled to start with ``I would''. For clarity, we observed that omitting a trailing space after this prefill produced more fluent responses. Lastly, all inventory and SJT responses were obtained under greedy decoding.

\clearpage

\section{OCEAN Injection Results for Llama-3.2-1B-Instruct}

\begin{figure}[!htbp]
    \centering
    \includegraphics[page=1]{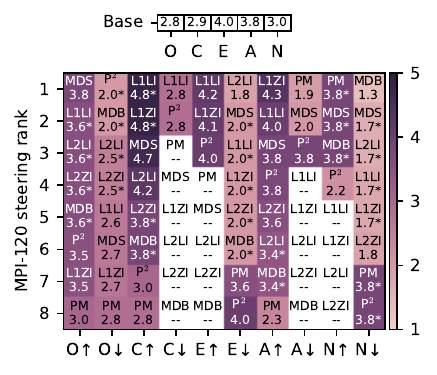}
    \includegraphics[page=2]{analysis/sweeps/Llama-3.2-1B-Instruct/heatplots.pdf}
    \caption{Ranking of steering methods on Llama-3.2-1B-Instruct by OCEAN trait and direction, and task. Based on each method’s best scores, with asterisks denoting ties in the unrounded results.}
\end{figure}

\begin{figure}[!htbp]
    \centering

    \begin{minipage}[t]{0.49\textwidth}
        \centering
        \includegraphics[page=1]{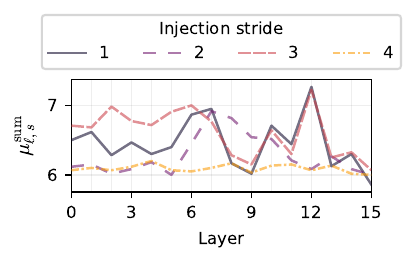}
    \end{minipage}
    \hfill
    \begin{minipage}[t]{0.49\textwidth}
        \centering
        \includegraphics[page=2]{analysis/sweeps/Llama-3.2-1B-Instruct/phi_plots.pdf}
    \end{minipage}
    \caption{Overall MDS injections steering performance on Llama-3.2-1B-Instruct by injection stride $s$ and model layer $\ell$. The line plot on the left shows MPI-120 results, and the shaded-area plot on the right shows SJT results.}
\end{figure}

\begin{figure}[!htbp]
    \centering
    \includegraphics[page=1]{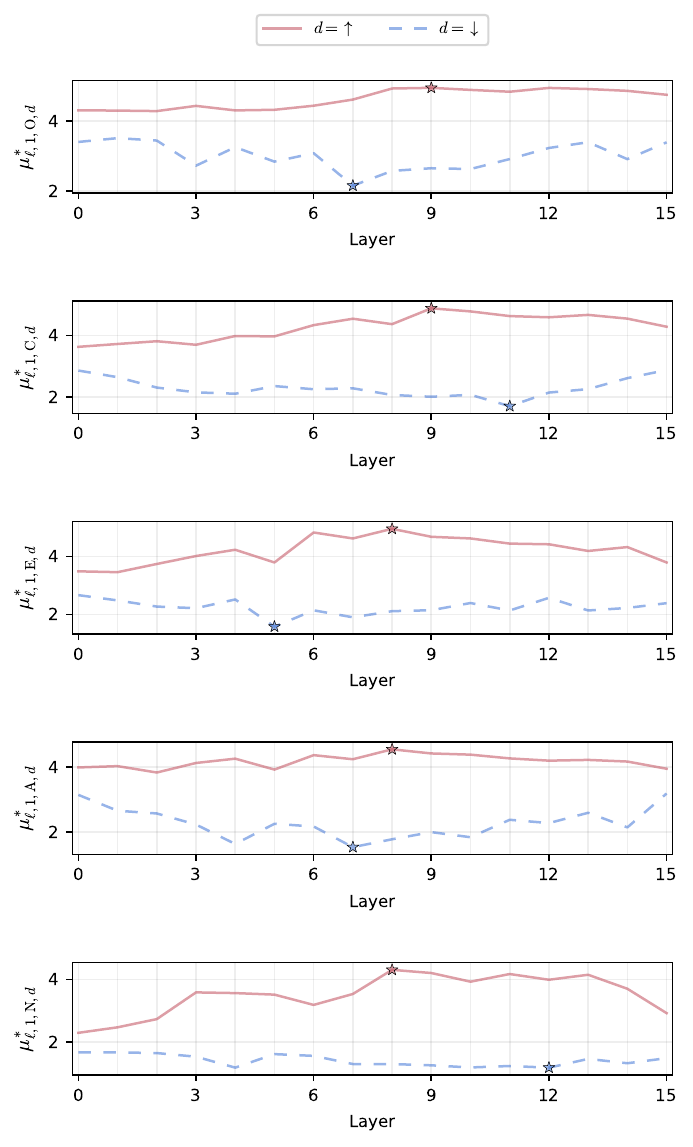}
    \caption{Layerwise extreme OCEAN steering scores on the SJTs task by direction $d \in \{\uparrow, \downarrow\}$ and model layer $\ell$, after applying MDS injections with injection stride $s = 1$ on Llama-3.2-1B-Instruct. Stars mark the strongest steering effects across layers ($\phi_{1,t,d}$).}
\end{figure}

\clearpage
\begin{figure}[!htbp]
	\centering
	\includegraphics{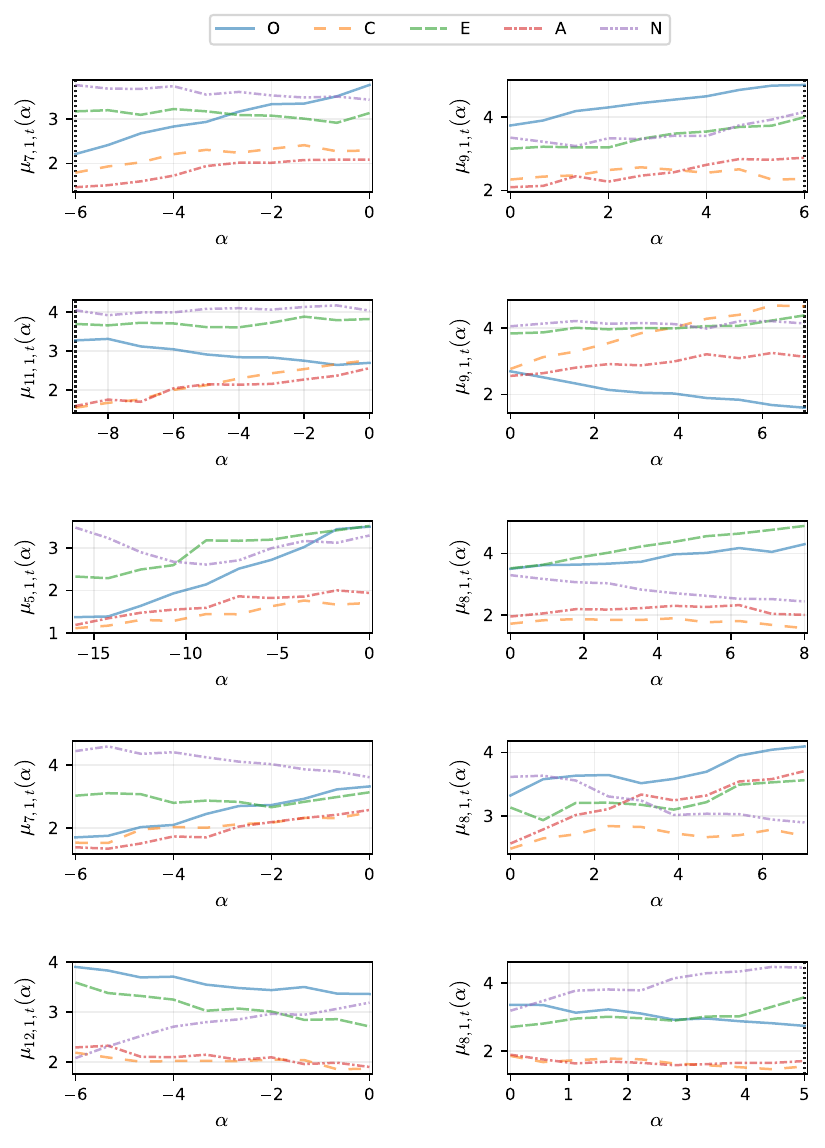}
	\caption{OCEAN scores for Llama-3.2-1B-Instruct on SJTs, under MDS injections with $s = 1$, using the best-performing layer $\ell$ for each trait-direction pair and 10 equidistant $\alpha$ values from 0 (no steering) to the best-performing $\alpha$.  From top to bottom, rows show openness, conscientiousness, extraversion, agreeableness, and neuroticism results. Negative $\alpha$ steers away from the target construct, and positive $\alpha$ steers toward it. Fluency was evaluated only in the responses to the corresponding SJTs. Vertical lines indicate some nonfluent SJT responses.}
\end{figure}

\clearpage

\section{OCEAN Injection Results for Llama-3.2-3B-Instruct}

\begin{figure}[!htbp]
    \centering
    \includegraphics[page=1]{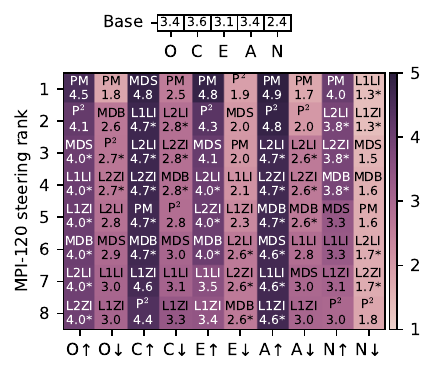}
    \includegraphics[page=2]{analysis/sweeps/Llama-3.2-3B-Instruct/heatplots.pdf}
    \caption{Ranking of steering methods on Llama-3.2-3B-Instruct by OCEAN trait and direction, and task. Based on each method’s best scores, with asterisks denoting ties in the unrounded results.}
\end{figure}

\begin{figure}[!htbp]
    \centering

    \begin{minipage}[t]{0.49\textwidth}
        \centering
        \includegraphics[page=1]{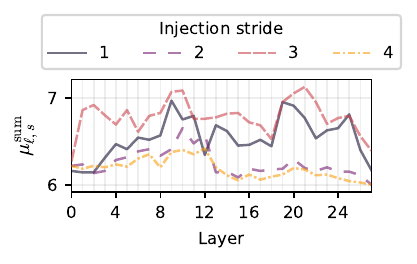}
    \end{minipage}
    \hfill
    \begin{minipage}[t]{0.49\textwidth}
        \centering
        \includegraphics[page=2]{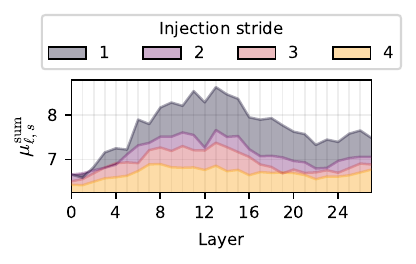}
    \end{minipage}
    \caption{Overall MDS injections steering performance on Llama-3.2-3B-Instruct by injection stride $s$ and model layer $\ell$. The line plot on the left shows MPI-120 results, and the shaded-area plot on the right shows SJT results.}
\end{figure}

\begin{figure}[!htbp]
    \centering
    \includegraphics[page=1]{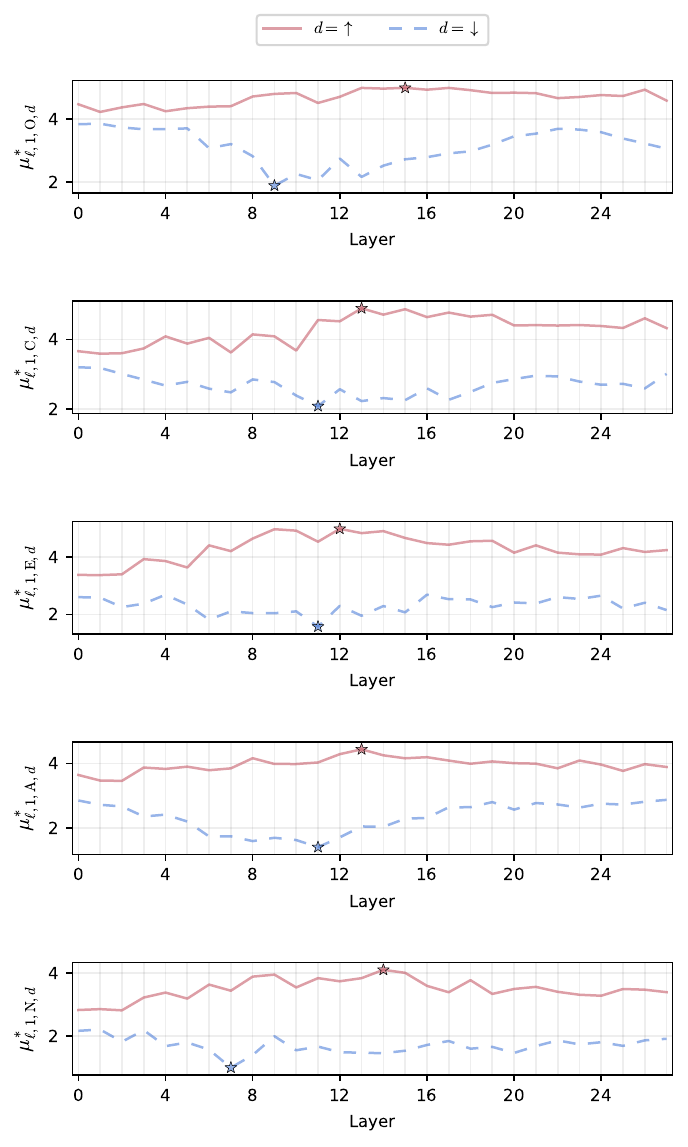}
    \caption{Layerwise extreme OCEAN steering scores on the SJTs task by direction $d \in \{\uparrow, \downarrow\}$ and model layer $\ell$, after applying MDS injections with injection stride $s = 1$ on Llama-3.2-3B-Instruct. Stars mark the strongest steering effects across layers ($\phi_{1,t,d}$).}
\end{figure}

\clearpage
\begin{figure}[!htbp]
	\centering
	\includegraphics{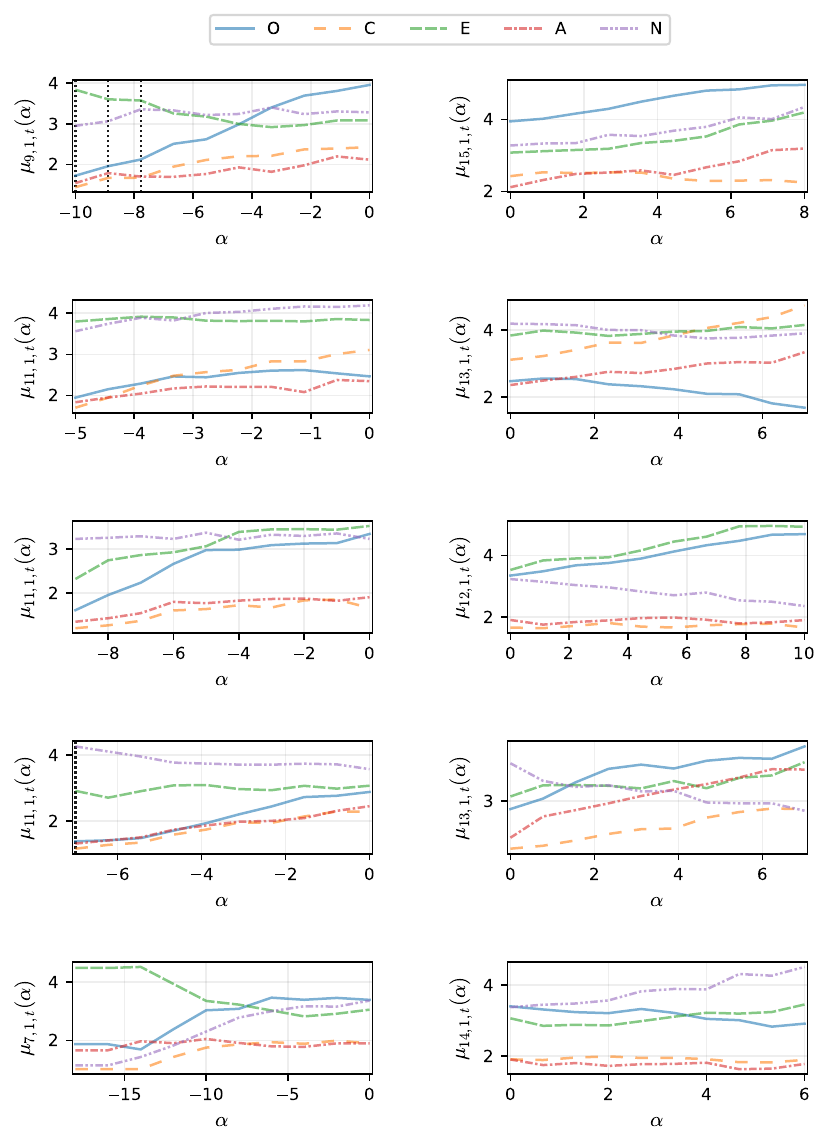}
	\caption{OCEAN scores for Llama-3.2-3B-Instruct on SJTs, under MDS injections with $s = 1$, using the best-performing layer $\ell$ for each trait-direction pair and 10 equidistant $\alpha$ values from 0 (no steering) to the best-performing $\alpha$.  From top to bottom, rows show openness, conscientiousness, extraversion, agreeableness, and neuroticism results. Negative $\alpha$ steers away from the target construct, and positive $\alpha$ steers toward it. Fluency was evaluated only in the responses to the corresponding SJTs. Vertical lines indicate some nonfluent SJT responses.}
\end{figure}

\clearpage

\section{OCEAN Injection Results for Llama-3.1-8B-Instruct}

\begin{figure}[!htbp]
    \centering
    \includegraphics[page=1]{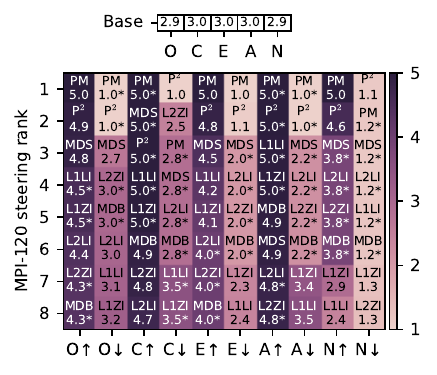}
    \includegraphics[page=2]{analysis/sweeps/Llama-3.1-8B-Instruct/heatplots.pdf}
    \caption{Ranking of steering methods on Llama-3.1-8B-Instruct by OCEAN trait and direction, and task. Based on each method’s best scores, with asterisks denoting ties in the unrounded results.}
\end{figure}

\begin{figure}[!htbp]
    \centering

    \begin{minipage}[t]{0.49\textwidth}
        \centering
        \includegraphics[page=1]{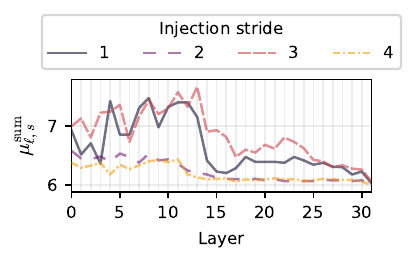}
    \end{minipage}
    \hfill
    \begin{minipage}[t]{0.49\textwidth}
        \centering
        \includegraphics[page=2]{analysis/sweeps/Llama-3.1-8B-Instruct/phi_plots.pdf}
    \end{minipage}
    \caption{Overall MDS injections steering performance on Llama-3.1-8B-Instruct by injection stride $s$ and model layer $\ell$. The line plot on the left shows MPI-120 results, and the shaded-area plot on the right shows SJT results.}
\end{figure}

\begin{figure}[!htbp]
    \centering
    \includegraphics[page=1]{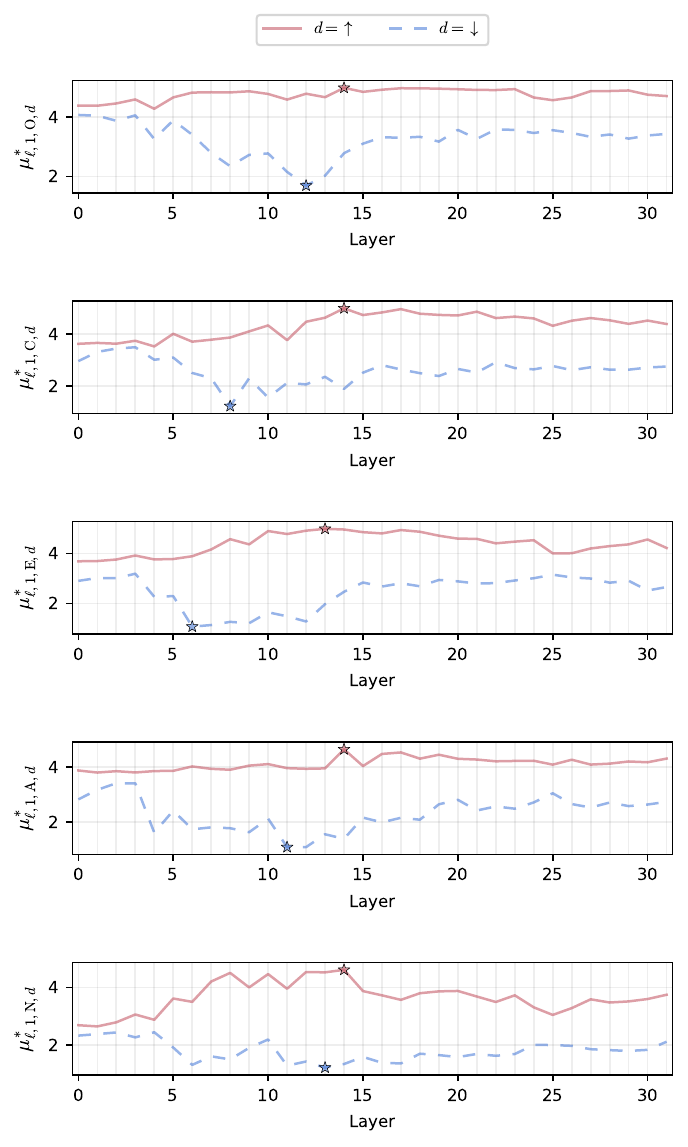}
    \caption{Layerwise extreme OCEAN steering scores on the SJTs task by direction $d \in \{\uparrow, \downarrow\}$ and model layer $\ell$, after applying MDS injections with injection stride $s = 1$ on Llama-3.1-8B-Instruct. Stars mark the strongest steering effects across layers ($\phi_{1,t,d}$).}
\end{figure}

\clearpage
\begin{figure}[!htbp]
	\centering
	\includegraphics{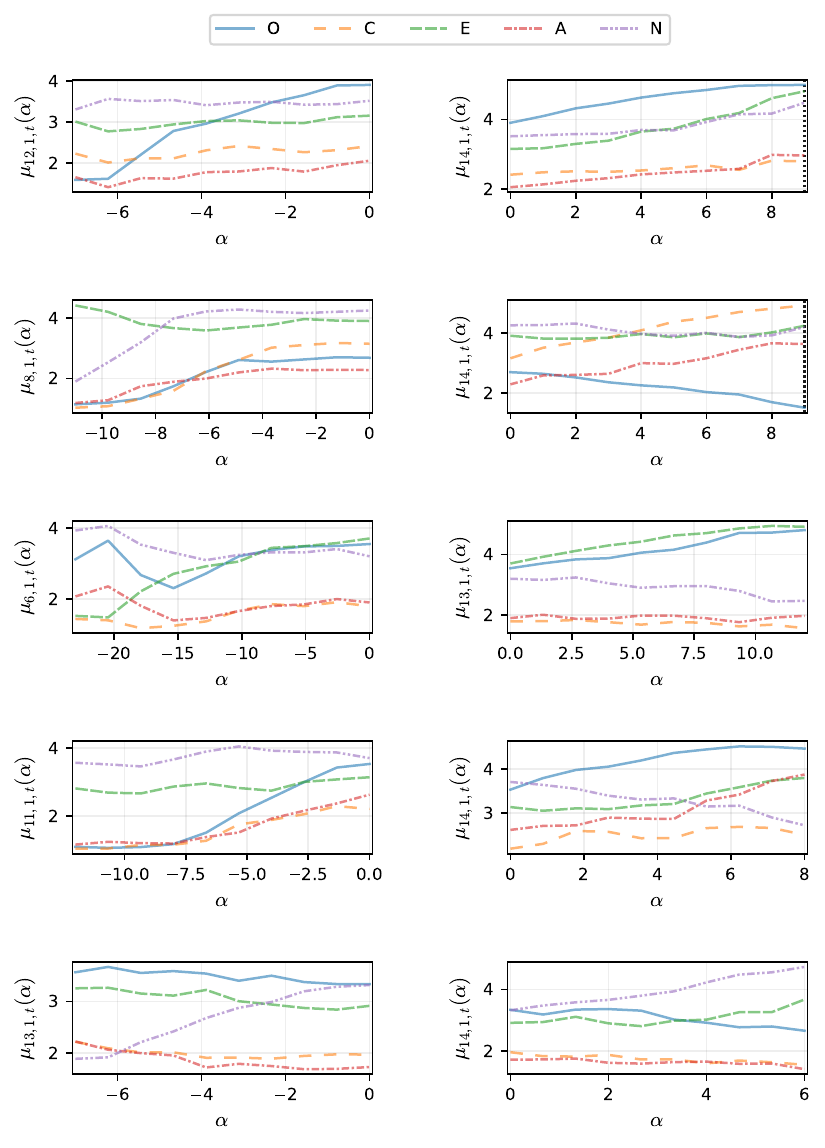}
	\caption{OCEAN scores for Llama-3.1-8B-Instruct on SJTs, under MDS injections with $s = 1$, using the best-performing layer $\ell$ for each trait-direction pair and 10 equidistant $\alpha$ values from 0 (no steering) to the best-performing $\alpha$.  From top to bottom, rows show openness, conscientiousness, extraversion, agreeableness, and neuroticism results. Negative $\alpha$ steers away from the target construct, and positive $\alpha$ steers toward it. Fluency was evaluated only in the responses to the corresponding SJTs. Vertical lines indicate some nonfluent SJT responses.}
\end{figure}

\clearpage

\section{OCEAN Injection Results for Qwen3-1.7B}

\begin{figure}[!htbp]
    \centering
    \includegraphics[page=1]{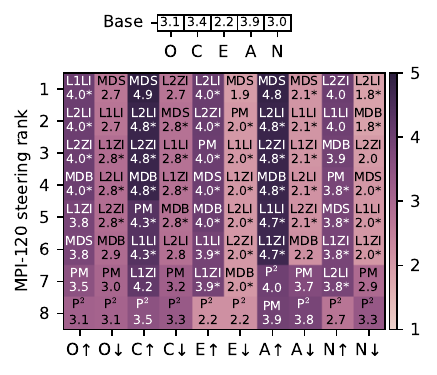}
    \includegraphics[page=2]{analysis/sweeps/Qwen3-1.7B/heatplots.pdf}
    \caption{Ranking of steering methods on Qwen3-1.7B by OCEAN trait and direction, and task. Based on each method’s best scores, with asterisks denoting ties in the unrounded results.}
\end{figure}

\begin{figure}[!htbp]
    \centering

    \begin{minipage}[t]{0.49\textwidth}
        \centering
        \includegraphics[page=1]{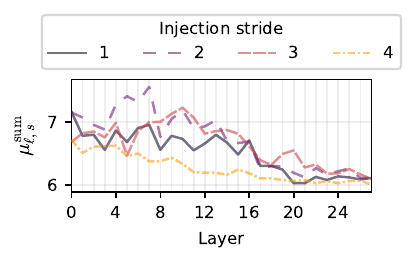}
    \end{minipage}
    \hfill
    \begin{minipage}[t]{0.49\textwidth}
        \centering
        \includegraphics[page=2]{analysis/sweeps/Qwen3-1.7B/phi_plots.pdf}
    \end{minipage}
    \caption{Overall MDS injections steering performance on Qwen3-1.7B by injection stride $s$ and model layer $\ell$. The line plot on the left shows MPI-120 results, and the shaded-area plot on the right shows SJT results.}
\end{figure}

\begin{figure}[!htbp]
    \centering
    \includegraphics[page=1]{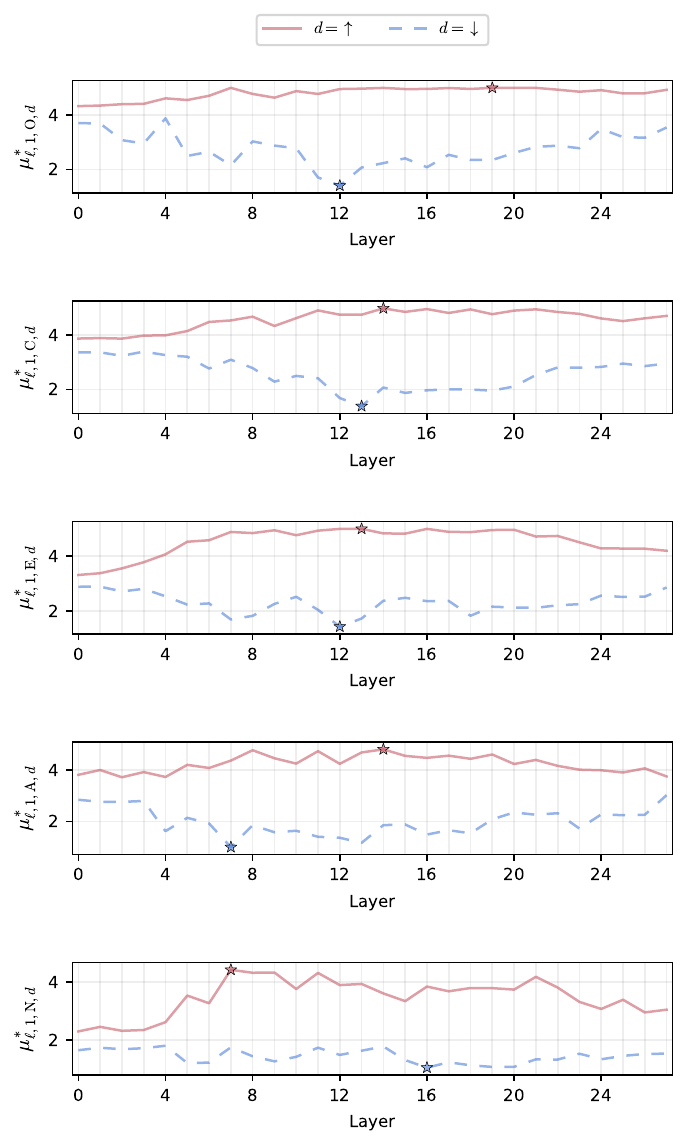}
    \caption{Layerwise extreme OCEAN steering scores on the SJTs task by direction $d \in \{\uparrow, \downarrow\}$ and model layer $\ell$, after applying MDS injections with injection stride $s = 1$ on Qwen3-1.7B. Stars mark the strongest steering effects across layers ($\phi_{1,t,d}$).}
\end{figure}

\clearpage
\begin{figure}[!htbp]
	\centering
	\includegraphics{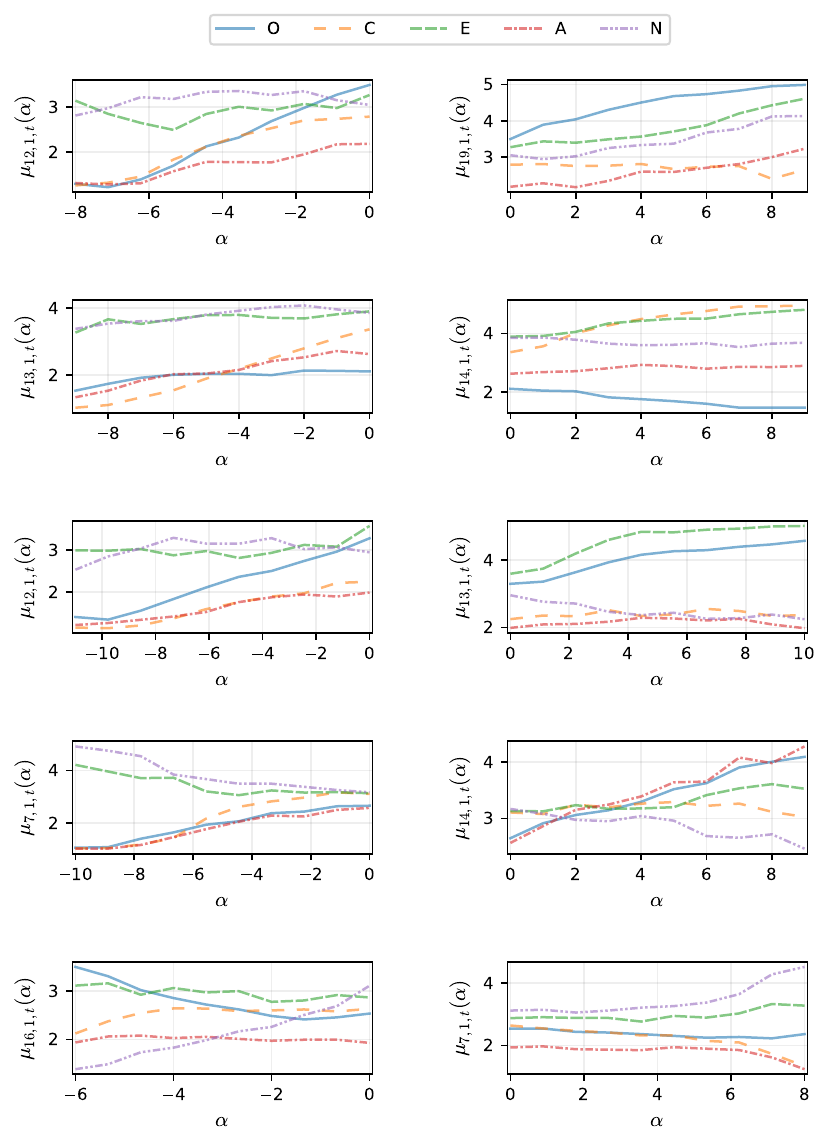}
	\caption{OCEAN scores for Qwen3-1.7B on SJTs, under MDS injections with $s = 1$, using the best-performing layer $\ell$ for each trait-direction pair and 10 equidistant $\alpha$ values from 0 (no steering) to the best-performing $\alpha$.  From top to bottom, rows show openness, conscientiousness, extraversion, agreeableness, and neuroticism results. Negative $\alpha$ steers away from the target construct, and positive $\alpha$ steers toward it. Fluency was evaluated only in the responses to the corresponding SJTs. Vertical lines indicate some nonfluent SJT responses.}
\end{figure}

\clearpage

\section{OCEAN Injection Results for Qwen3-4B}

\begin{figure}[!htbp]
    \centering
    \includegraphics[page=1]{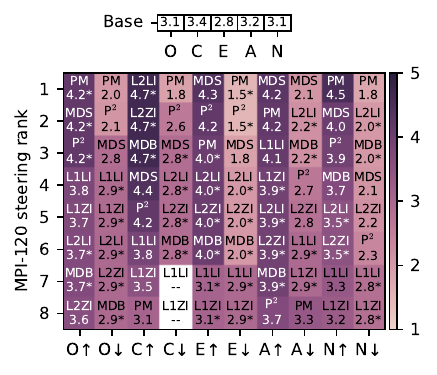}
    \includegraphics[page=2]{analysis/sweeps/Qwen3-4B/heatplots.pdf}
    \caption{Ranking of steering methods on Qwen3-4B by OCEAN trait and direction, and task. Based on each method’s best scores, with asterisks denoting ties in the unrounded results.}
\end{figure}

\begin{figure}[!htbp]
    \centering

    \begin{minipage}[t]{0.49\textwidth}
        \centering
        \includegraphics[page=1]{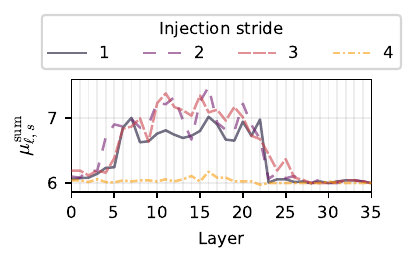}
    \end{minipage}
    \hfill
    \begin{minipage}[t]{0.49\textwidth}
        \centering
        \includegraphics[page=2]{analysis/sweeps/Qwen3-4B/phi_plots.pdf}
    \end{minipage}
    \caption{Overall MDS injections steering performance on Qwen3-4B by injection stride $s$ and model layer $\ell$. The line plot on the left shows MPI-120 results, and the shaded-area plot on the right shows SJT results.}
\end{figure}

\begin{figure}[!htbp]
    \centering
    \includegraphics[page=1]{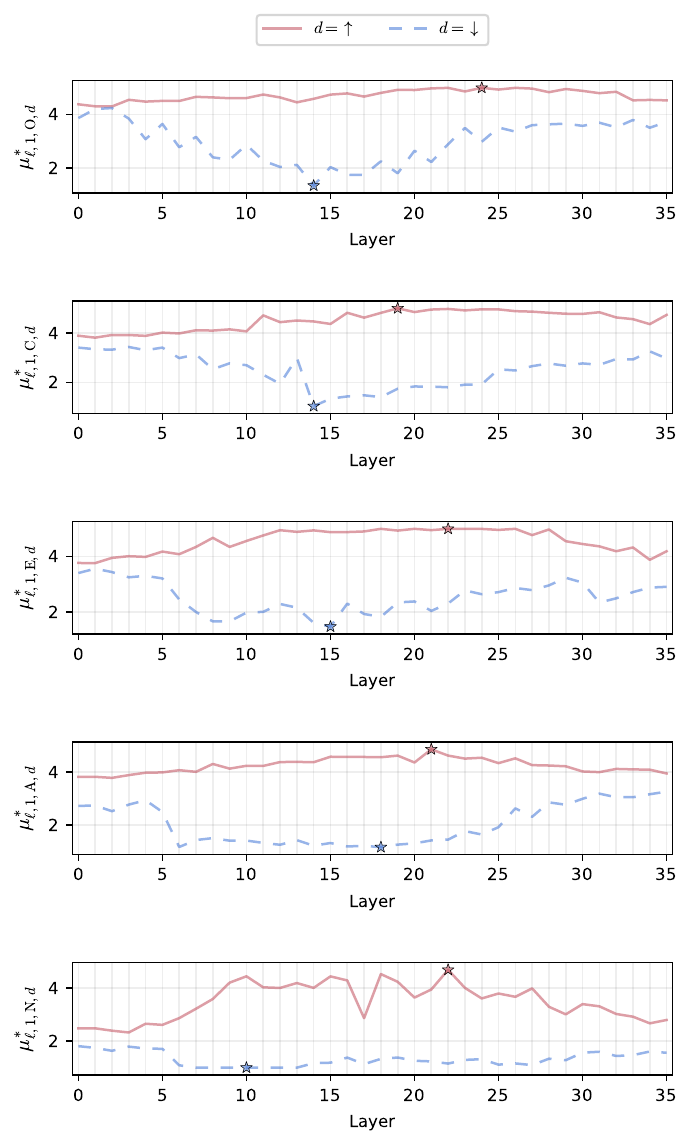}
    \caption{Layerwise extreme OCEAN steering scores on the SJTs task by direction $d \in \{\uparrow, \downarrow\}$ and model layer $\ell$, after applying MDS injections with injection stride $s = 1$ on Qwen3-4B. Stars mark the strongest steering effects across layers ($\phi_{1,t,d}$).}
\end{figure}

\clearpage
\begin{figure}[!htbp]
	\centering
	\includegraphics{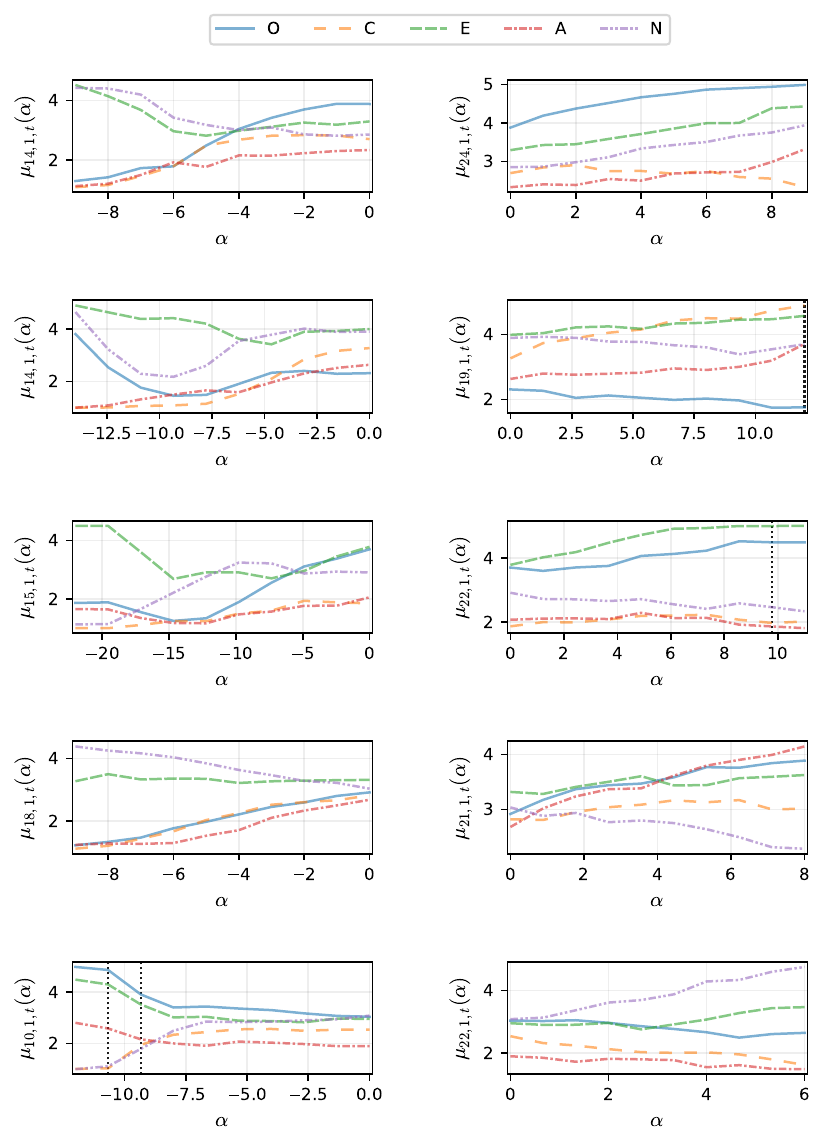}
	\caption{OCEAN scores for Qwen3-4B on SJTs, under MDS injections with $s = 1$, using the best-performing layer $\ell$ for each trait-direction pair and 10 equidistant $\alpha$ values from 0 (no steering) to the best-performing $\alpha$.  From top to bottom, rows show openness, conscientiousness, extraversion, agreeableness, and neuroticism results. Negative $\alpha$ steers away from the target construct, and positive $\alpha$ steers toward it. Fluency was evaluated only in the responses to the corresponding SJTs. Vertical lines indicate some nonfluent SJT responses.}
\end{figure}

\clearpage

\section{OCEAN Injection Results for Qwen3-8B}

\begin{figure}[!htbp]
    \centering
    \includegraphics[page=1]{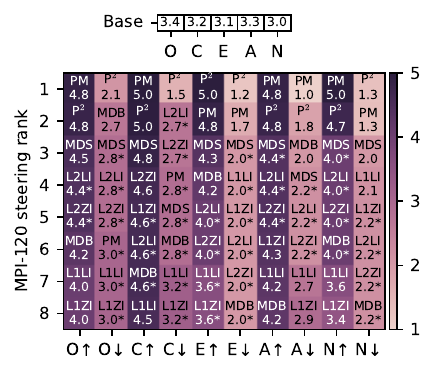}
    \includegraphics[page=2]{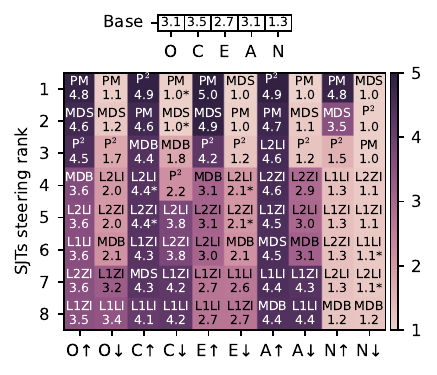}
    \caption{Ranking of steering methods on Qwen3-8B by OCEAN trait and direction, and task. Based on each method’s best scores, with asterisks denoting ties in the unrounded results.}
\end{figure}

\begin{figure}[!htbp]
    \centering

    \begin{minipage}[t]{0.49\textwidth}
        \centering
        \includegraphics[page=1]{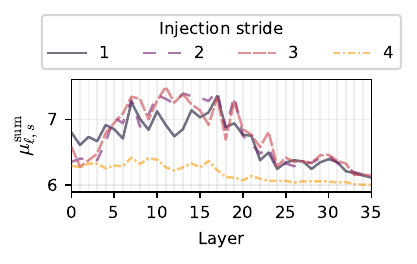}
    \end{minipage}
    \hfill
    \begin{minipage}[t]{0.49\textwidth}
        \centering
        \includegraphics[page=2]{analysis/sweeps/Qwen3-8B/phi_plots.pdf}
    \end{minipage}
    \caption{Overall MDS injections steering performance on Qwen3-8B by injection stride $s$ and model layer $\ell$. The line plot on the left shows MPI-120 results, and the shaded-area plot on the right shows SJT results.}
\end{figure}

\begin{figure}[!htbp]
    \centering
    \includegraphics[page=1]{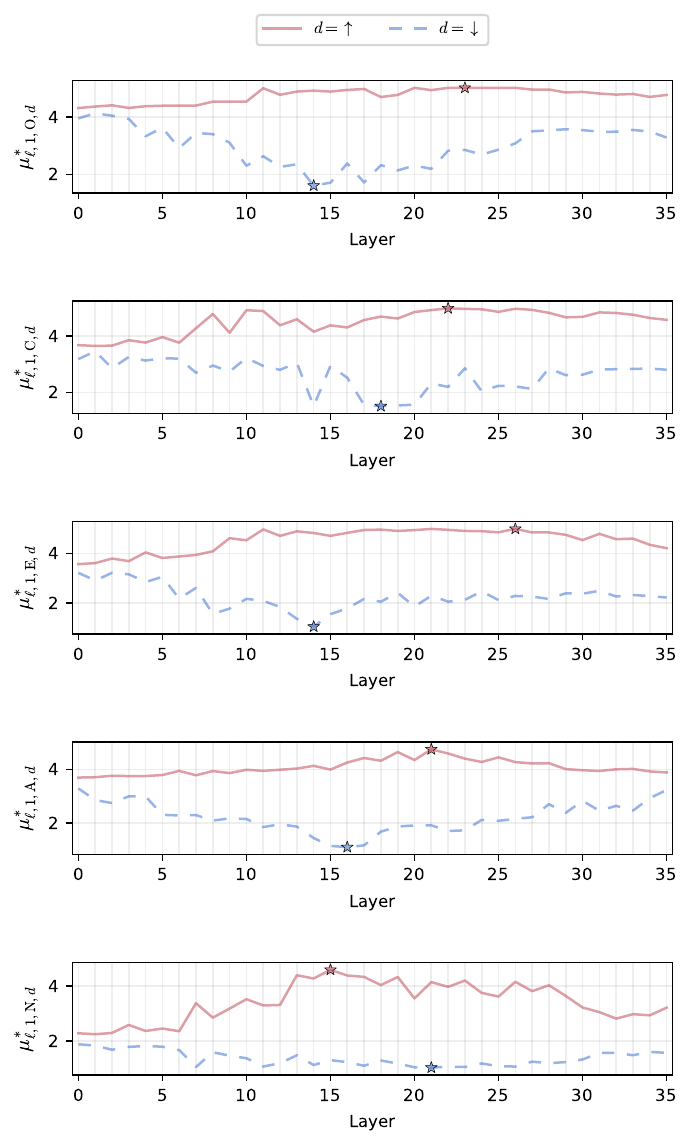}
    \caption{Layerwise extreme OCEAN steering scores on the SJTs task by direction $d \in \{\uparrow, \downarrow\}$ and model layer $\ell$, after applying MDS injections with injection stride $s = 1$ on Qwen3-8B. Stars mark the strongest steering effects across layers ($\phi_{1,t,d}$).}
\end{figure}

\clearpage
\begin{figure}[!htbp]
	\centering
	\includegraphics{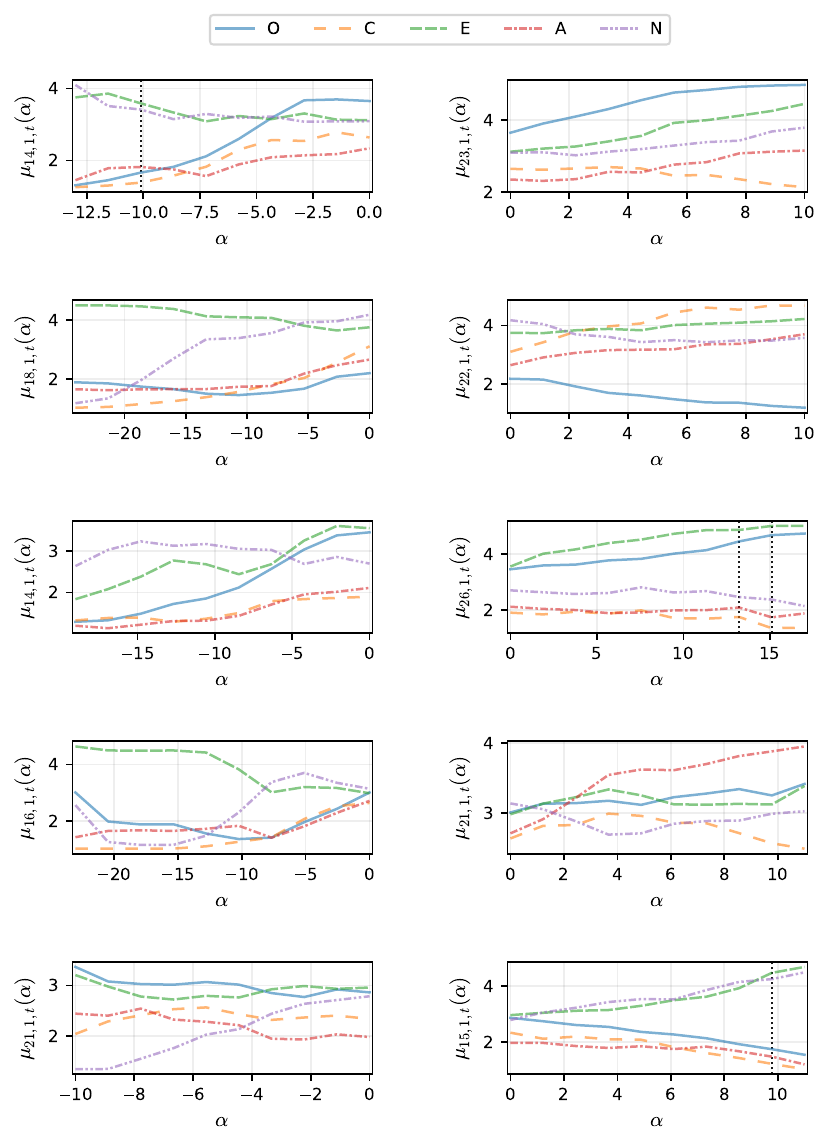}
	\caption{OCEAN scores for Qwen3-8B on SJTs, under MDS injections with $s = 1$, using the best-performing layer $\ell$ for each trait-direction pair and 10 equidistant $\alpha$ values from 0 (no steering) to the best-performing $\alpha$.  From top to bottom, rows show openness, conscientiousness, extraversion, agreeableness, and neuroticism results. Negative $\alpha$ steers away from the target construct, and positive $\alpha$ steers toward it. Fluency was evaluated only in the responses to the corresponding SJTs. Vertical lines indicate some nonfluent SJT responses.}
\end{figure}

\clearpage

\section{OCEAN Injection Results for Qwen3-14B}

\begin{figure}[!htbp]
    \centering
    \includegraphics[page=1]{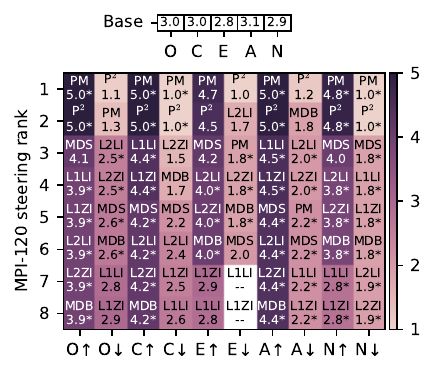}
    \includegraphics[page=2]{analysis/sweeps/Qwen3-14B/heatplots.pdf}
    \caption{Ranking of steering methods on Qwen3-14B by OCEAN trait and direction, and task. Based on each method’s best scores, with asterisks denoting ties in the unrounded results.}
\end{figure}

\begin{figure}[!htbp]
    \centering

    \begin{minipage}[t]{0.49\textwidth}
        \centering
        \includegraphics[page=1]{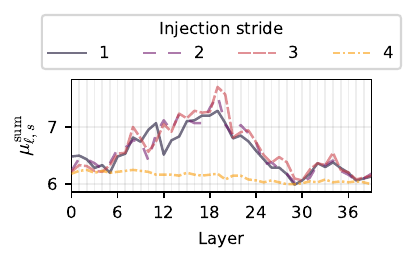}
    \end{minipage}
    \hfill
    \begin{minipage}[t]{0.49\textwidth}
        \centering
        \includegraphics[page=2]{analysis/sweeps/Qwen3-14B/phi_plots.pdf}
    \end{minipage}
    \caption{Overall MDS injections steering performance on Qwen3-14B by injection stride $s$ and model layer $\ell$. The line plot on the left shows MPI-120 results, and the shaded-area plot on the right shows SJT results.}
\end{figure}

\begin{figure}[!htbp]
    \centering
    \includegraphics[page=1]{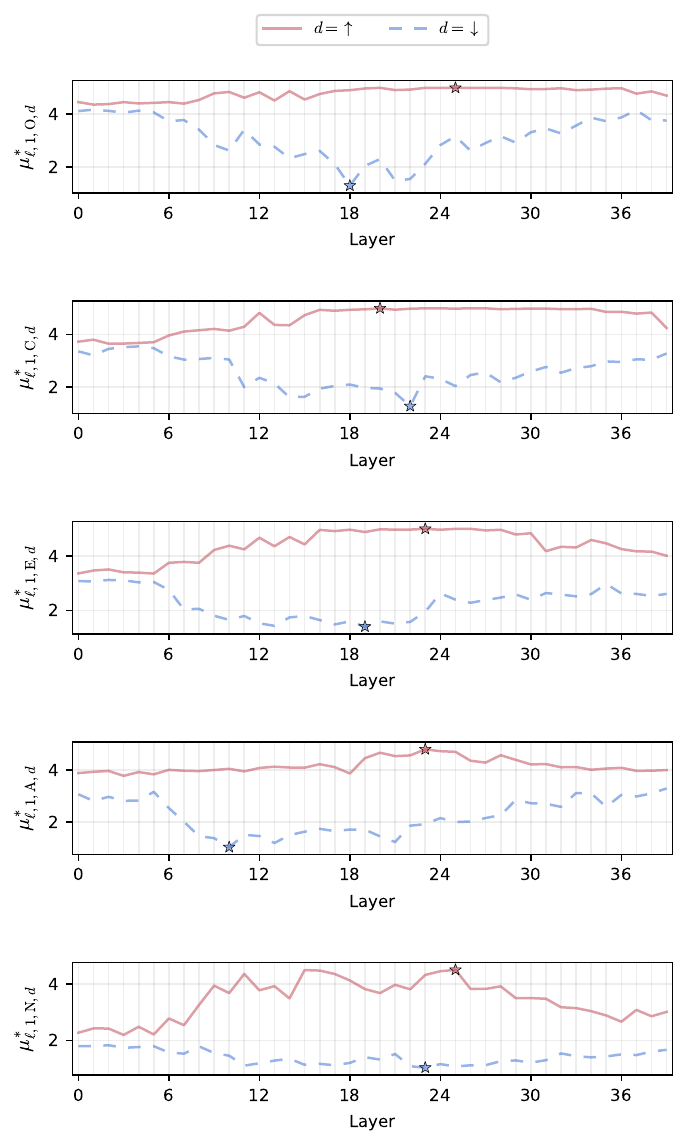}
    \caption{Layerwise extreme OCEAN steering scores on the SJTs task by direction $d \in \{\uparrow, \downarrow\}$ and model layer $\ell$, after applying MDS injections with injection stride $s = 1$ on Qwen3-14B. Stars mark the strongest steering effects across layers ($\phi_{1,t,d}$).}
\end{figure}

\clearpage
\begin{figure}[!htbp]
	\centering
	\includegraphics{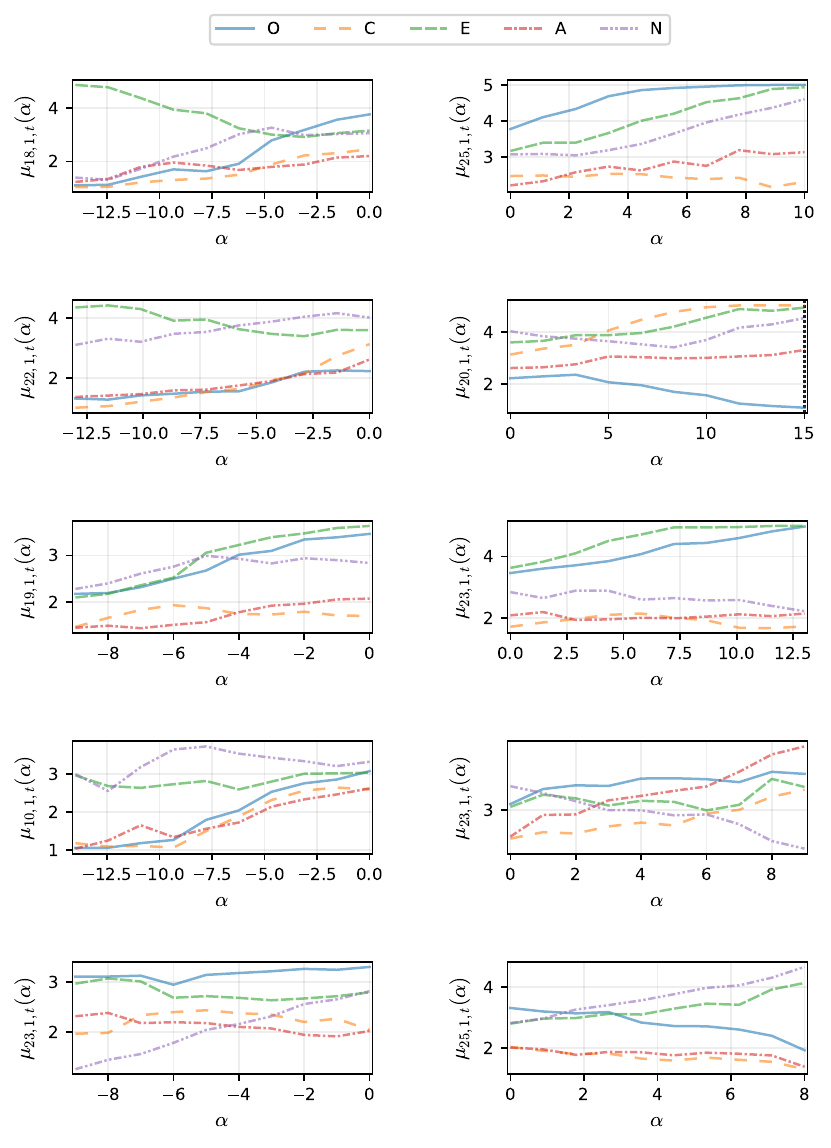}
	\caption{OCEAN scores for Qwen3-14B on SJTs, under MDS injections with $s = 1$, using the best-performing layer $\ell$ for each trait-direction pair and 10 equidistant $\alpha$ values from 0 (no steering) to the best-performing $\alpha$.  From top to bottom, rows show openness, conscientiousness, extraversion, agreeableness, and neuroticism results. Negative $\alpha$ steers away from the target construct, and positive $\alpha$ steers toward it. Fluency was evaluated only in the responses to the corresponding SJTs. Vertical lines indicate some nonfluent SJT responses.}
\end{figure}

\clearpage

\section{OCEAN Injection Results for Qwen3-32B}

\begin{figure}[!htbp]
    \centering
    \includegraphics[page=1]{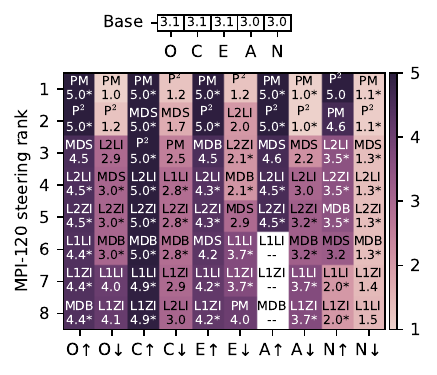}
    \includegraphics[page=2]{analysis/sweeps/Qwen3-32B/heatplots.pdf}
    \caption{Ranking of steering methods on Qwen3-32B by OCEAN trait and direction, and task. Based on each method’s best scores, with asterisks denoting ties in the unrounded results.}
\end{figure}

\begin{figure}[!htbp]
    \centering

    \begin{minipage}[t]{0.49\textwidth}
        \centering
        \includegraphics[page=1]{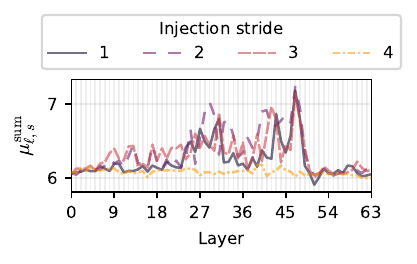}
    \end{minipage}
    \hfill
    \begin{minipage}[t]{0.49\textwidth}
        \centering
        \includegraphics[page=2]{analysis/sweeps/Qwen3-32B/phi_plots.pdf}
    \end{minipage}
    \caption{Overall MDS injections steering performance on Qwen3-32B by injection stride $s$ and model layer $\ell$. The line plot on the left shows MPI-120 results, and the shaded-area plot on the right shows SJT results.}
\end{figure}

\begin{figure}[!htbp]
    \centering
    \includegraphics[page=1]{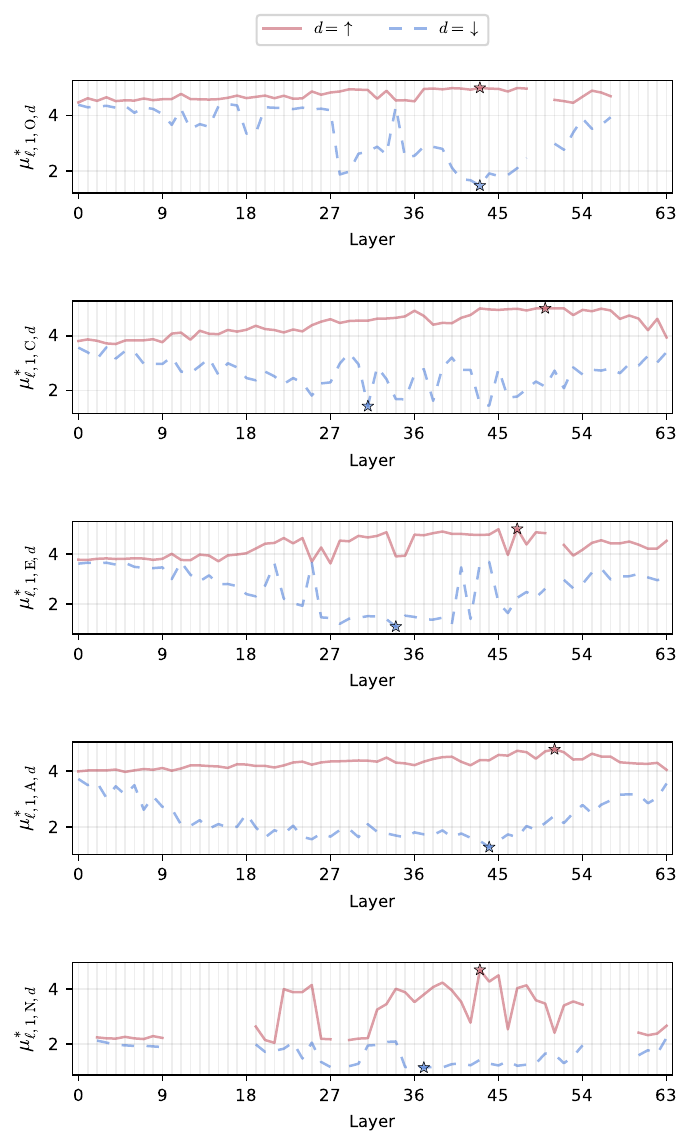}
    \caption{Layerwise extreme OCEAN steering scores on the SJTs task by direction $d \in \{\uparrow, \downarrow\}$ and model layer $\ell$, after applying MDS injections with injection stride $s = 1$ on Qwen3-32B. Stars mark the strongest steering effects across layers ($\phi_{1,t,d}$).}
\end{figure}

\clearpage
\begin{figure}[!htbp]
	\centering
	\includegraphics{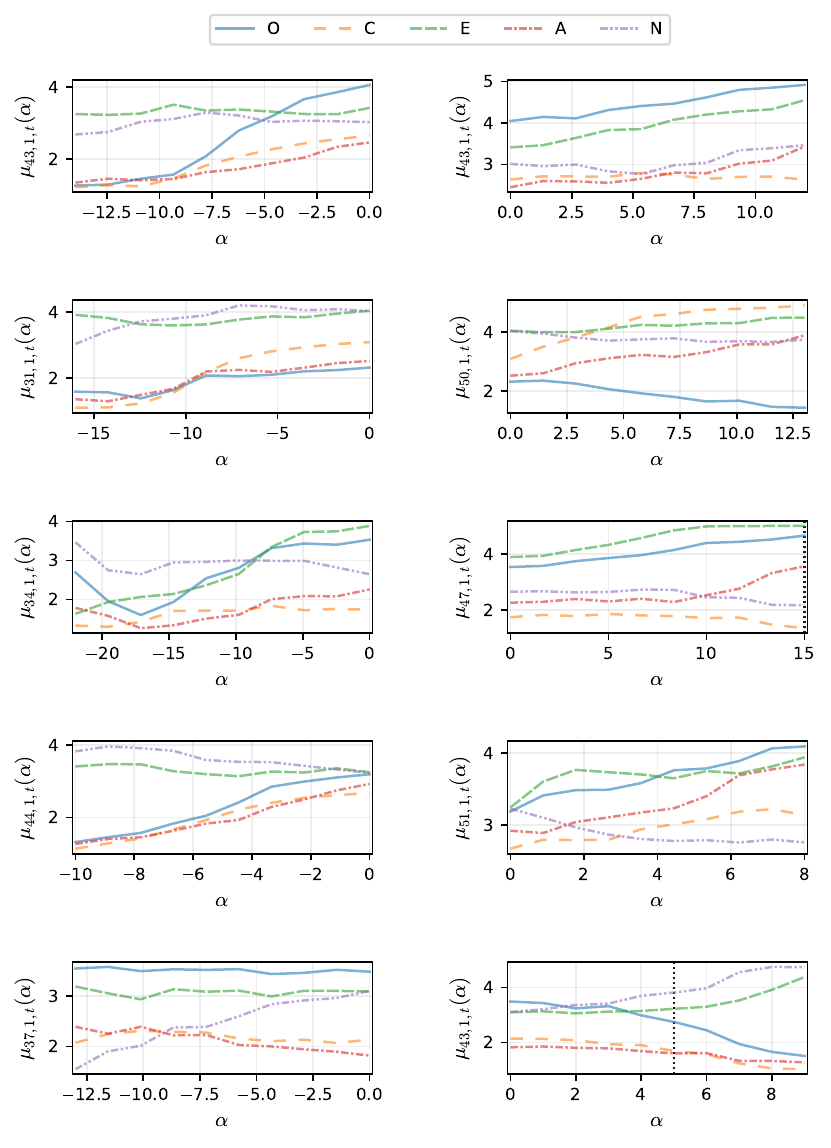}
	\caption{OCEAN scores for Qwen3-32B on SJTs, under MDS injections with $s = 1$, using the best-performing layer $\ell$ for each trait-direction pair and 10 equidistant $\alpha$ values from 0 (no steering) to the best-performing $\alpha$.  From top to bottom, rows show openness, conscientiousness, extraversion, agreeableness, and neuroticism results. Negative $\alpha$ steers away from the target construct, and positive $\alpha$ steers toward it. Fluency was evaluated only in the responses to the corresponding SJTs. Vertical lines indicate some nonfluent SJT responses.}
\end{figure}

\clearpage

\section{OCEAN Injection Results for gemma-3-1b-it}

\begin{figure}[!htbp]
    \centering
    \includegraphics[page=1]{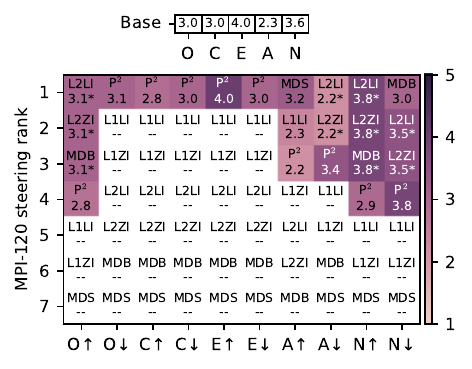}
    \includegraphics[page=2]{analysis/sweeps/gemma-3-1b-it/heatplots.pdf}
    \caption{Ranking of steering methods on gemma-3-1b-it by OCEAN trait and direction, and task. Based on each method’s best scores, with asterisks denoting ties in the unrounded results.}
\end{figure}

	No other valid MDS injection results to plot.

\clearpage

\section{OCEAN Injection Results for gemma-3-4b-it}

\begin{figure}[!htbp]
    \centering
    \includegraphics[page=1]{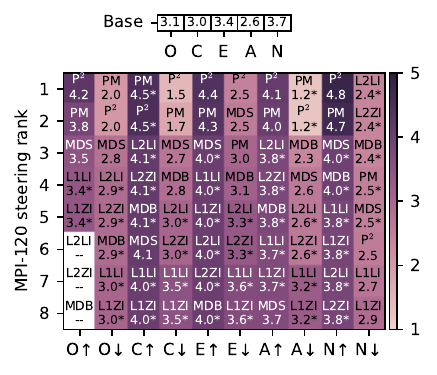}
    \includegraphics[page=2]{analysis/sweeps/gemma-3-4b-it/heatplots.pdf}
    \caption{Ranking of steering methods on gemma-3-4b-it by OCEAN trait and direction, and task. Based on each method’s best scores, with asterisks denoting ties in the unrounded results.}
\end{figure}

\begin{figure}[!htbp]
    \centering

    \begin{minipage}[t]{0.49\textwidth}
        \centering
        \includegraphics[page=1]{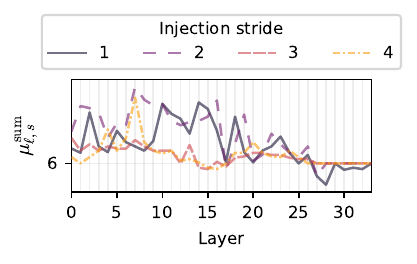}
    \end{minipage}
    \hfill
    \begin{minipage}[t]{0.49\textwidth}
        \centering
        \includegraphics[page=2]{analysis/sweeps/gemma-3-4b-it/phi_plots.pdf}
    \end{minipage}
    \caption{Overall MDS injections steering performance on gemma-3-4b-it by injection stride $s$ and model layer $\ell$. The line plot on the left shows MPI-120 results, and the shaded-area plot on the right shows SJT results.}
\end{figure}

\begin{figure}[!htbp]
    \centering
    \includegraphics[page=1]{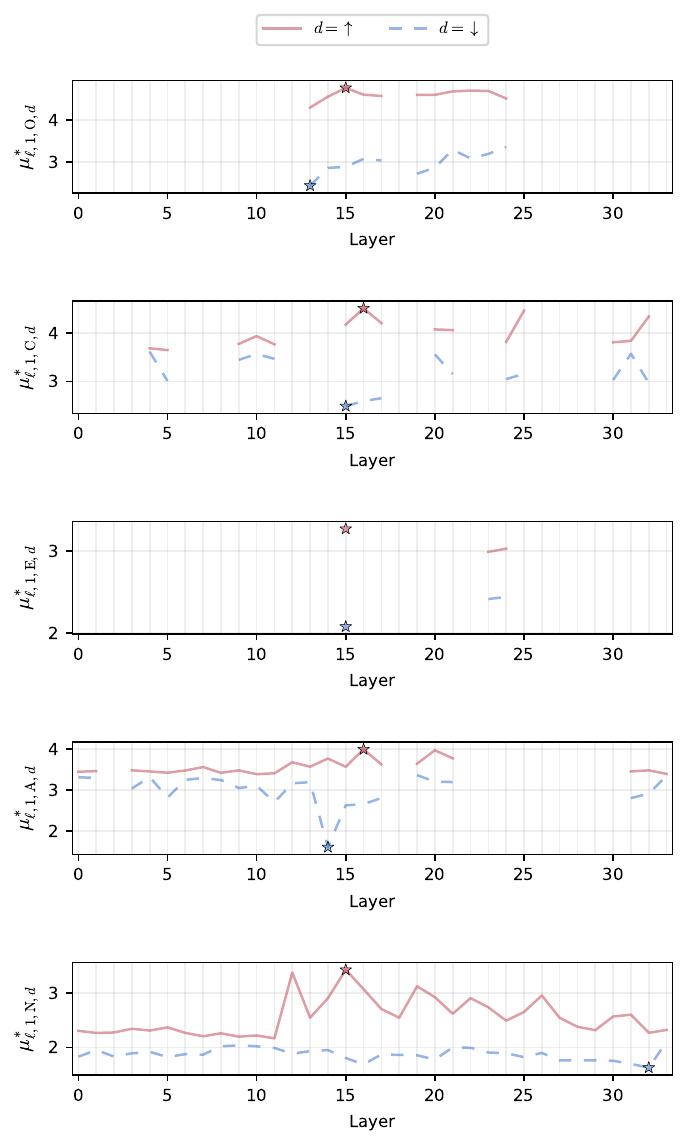}
    \caption{Layerwise extreme OCEAN steering scores on the SJTs task by direction $d \in \{\uparrow, \downarrow\}$ and model layer $\ell$, after applying MDS injections with injection stride $s = 1$ on gemma-3-4b-it. Stars mark the strongest steering effects across layers ($\phi_{1,t,d}$).}
\end{figure}

\clearpage
\begin{figure}[!htbp]
	\centering
	\includegraphics{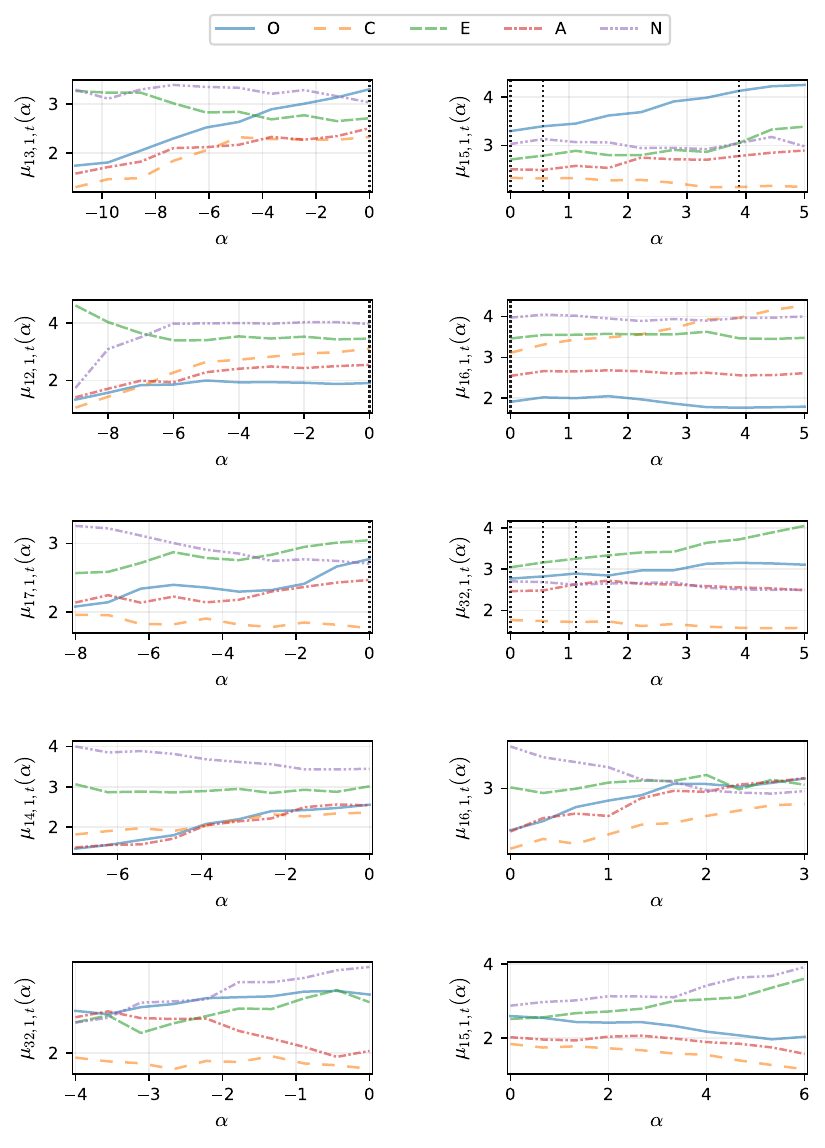}
	\caption{OCEAN scores for gemma-3-4b-it on SJTs, under MDS injections with $s = 1$, using the best-performing layer $\ell$ for each trait-direction pair and 10 equidistant $\alpha$ values from 0 (no steering) to the best-performing $\alpha$.  From top to bottom, rows show openness, conscientiousness, extraversion, agreeableness, and neuroticism results. Negative $\alpha$ steers away from the target construct, and positive $\alpha$ steers toward it. Fluency was evaluated only in the responses to the corresponding SJTs. Vertical lines indicate some nonfluent SJT responses.}
\end{figure}

\clearpage

\section{OCEAN Injection Results for gemma-3-12b-it}

\begin{figure}[!htbp]
    \centering
    \includegraphics[page=1]{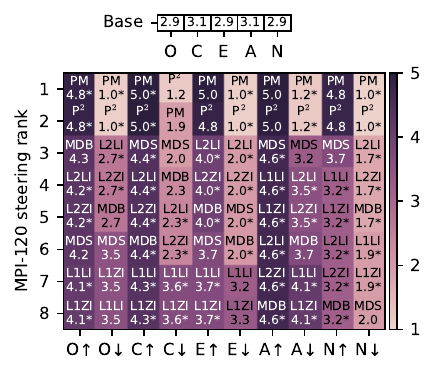}
    \includegraphics[page=2]{analysis/sweeps/gemma-3-12b-it/heatplots.pdf}
    \caption{Ranking of steering methods on gemma-3-12b-it by OCEAN trait and direction, and task. Based on each method’s best scores, with asterisks denoting ties in the unrounded results.}
\end{figure}

\begin{figure}[!htbp]
    \centering

    \begin{minipage}[t]{0.49\textwidth}
        \centering
        \includegraphics[page=1]{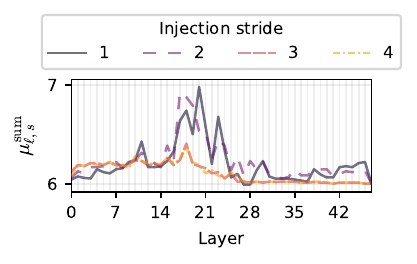}
    \end{minipage}
    \hfill
    \begin{minipage}[t]{0.49\textwidth}
        \centering
        \includegraphics[page=2]{analysis/sweeps/gemma-3-12b-it/phi_plots.pdf}
    \end{minipage}
    \caption{Overall MDS injections steering performance on gemma-3-12b-it by injection stride $s$ and model layer $\ell$. The line plot on the left shows MPI-120 results, and the shaded-area plot on the right shows SJT results.}
\end{figure}

\begin{figure}[!htbp]
    \centering
    \includegraphics[page=1]{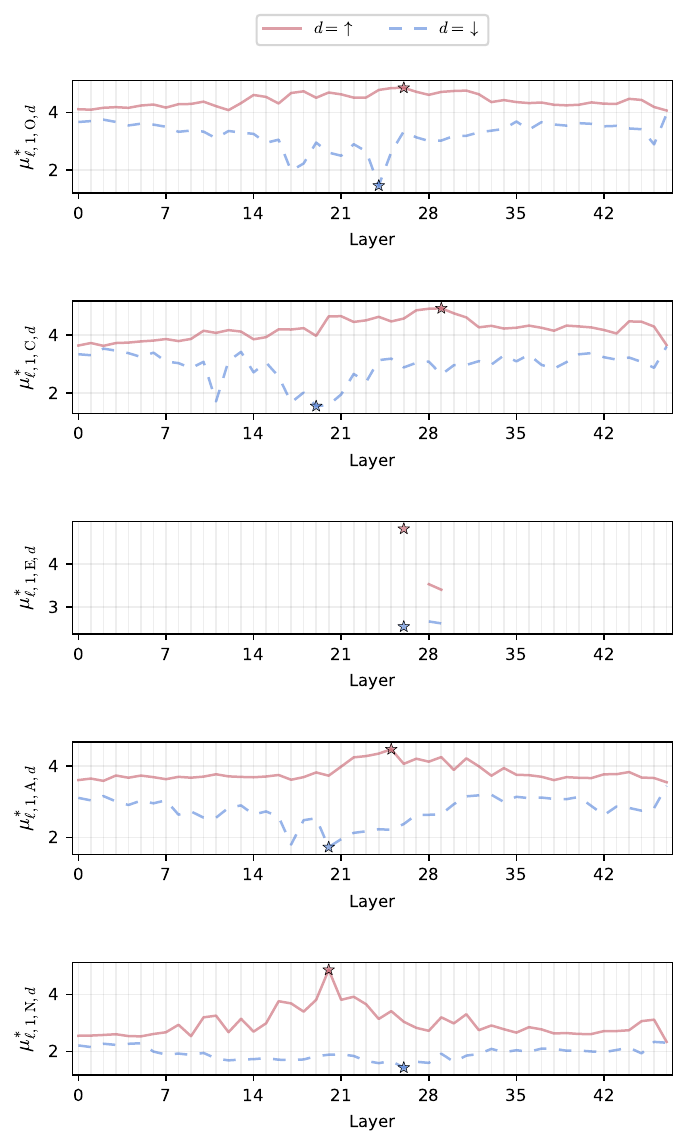}
    \caption{Layerwise extreme OCEAN steering scores on the SJTs task by direction $d \in \{\uparrow, \downarrow\}$ and model layer $\ell$, after applying MDS injections with injection stride $s = 1$ on gemma-3-12b-it. Stars mark the strongest steering effects across layers ($\phi_{1,t,d}$).}
\end{figure}

\clearpage
\begin{figure}[!htbp]
	\centering
	\includegraphics{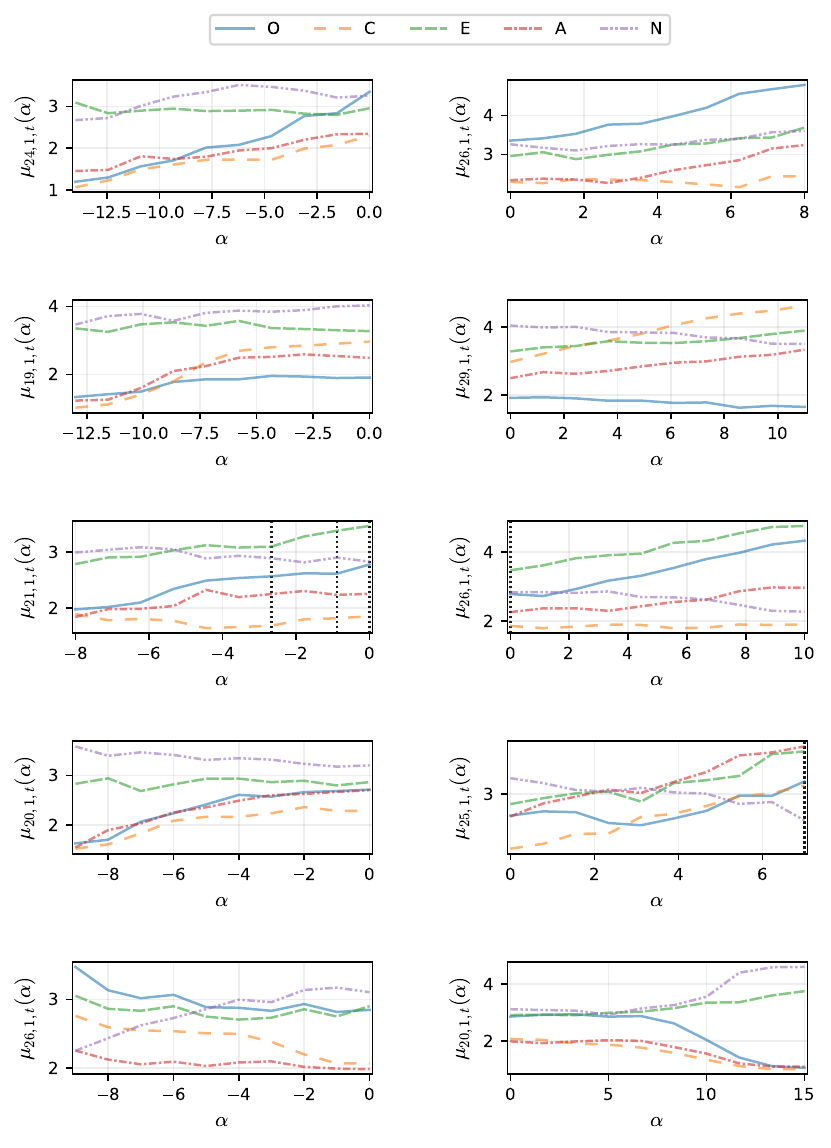}
	\caption{OCEAN scores for gemma-3-12b-it on SJTs, under MDS injections with $s = 1$, using the best-performing layer $\ell$ for each trait-direction pair and 10 equidistant $\alpha$ values from 0 (no steering) to the best-performing $\alpha$.  From top to bottom, rows show openness, conscientiousness, extraversion, agreeableness, and neuroticism results. Negative $\alpha$ steers away from the target construct, and positive $\alpha$ steers toward it. Fluency was evaluated only in the responses to the corresponding SJTs. Vertical lines indicate some nonfluent SJT responses.}
\end{figure}

\clearpage

\section{OCEAN Injection Results for gemma-3-27b-it}

\begin{figure}[!htbp]
    \centering
    \includegraphics[page=1]{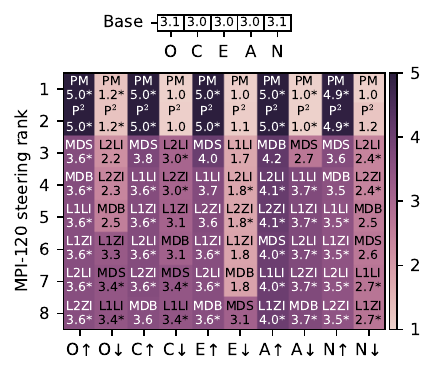}
    \includegraphics[page=2]{analysis/sweeps/gemma-3-27b-it/heatplots.pdf}
    \caption{Ranking of steering methods on gemma-3-27b-it by OCEAN trait and direction, and task. Based on each method’s best scores, with asterisks denoting ties in the unrounded results.}
\end{figure}

\begin{figure}[!htbp]
    \centering

    \begin{minipage}[t]{0.49\textwidth}
        \centering
        \includegraphics[page=1]{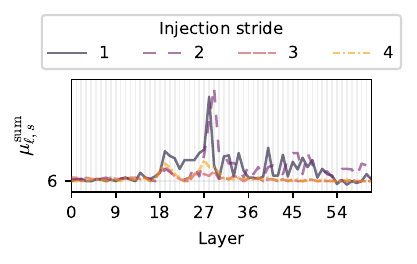}
    \end{minipage}
    \hfill
    \begin{minipage}[t]{0.49\textwidth}
        \centering
        \includegraphics[page=2]{analysis/sweeps/gemma-3-27b-it/phi_plots.pdf}
    \end{minipage}
    \caption{Overall MDS injections steering performance on gemma-3-27b-it by injection stride $s$ and model layer $\ell$. The line plot on the left shows MPI-120 results, and the shaded-area plot on the right shows SJT results.}
\end{figure}

\begin{figure}[!htbp]
    \centering
    \includegraphics[page=1]{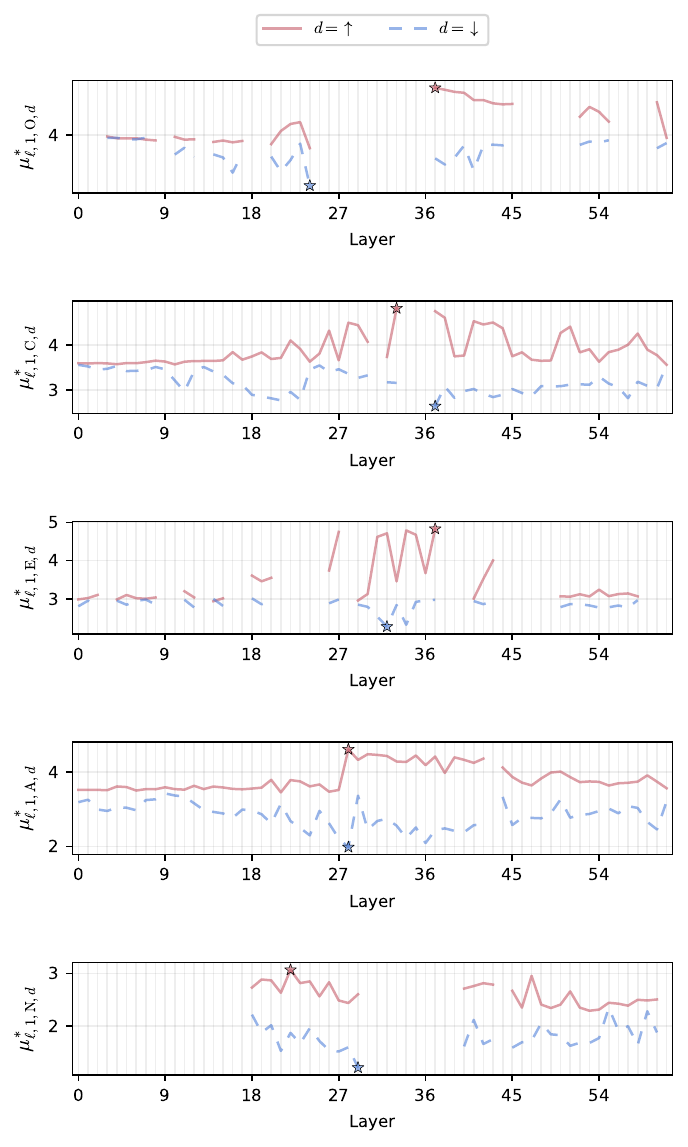}
    \caption{Layerwise extreme OCEAN steering scores on the SJTs task by direction $d \in \{\uparrow, \downarrow\}$ and model layer $\ell$, after applying MDS injections with injection stride $s = 1$ on gemma-3-27b-it. Stars mark the strongest steering effects across layers ($\phi_{1,t,d}$).}
\end{figure}

\clearpage
\begin{figure}[!htbp]
	\centering
	\includegraphics{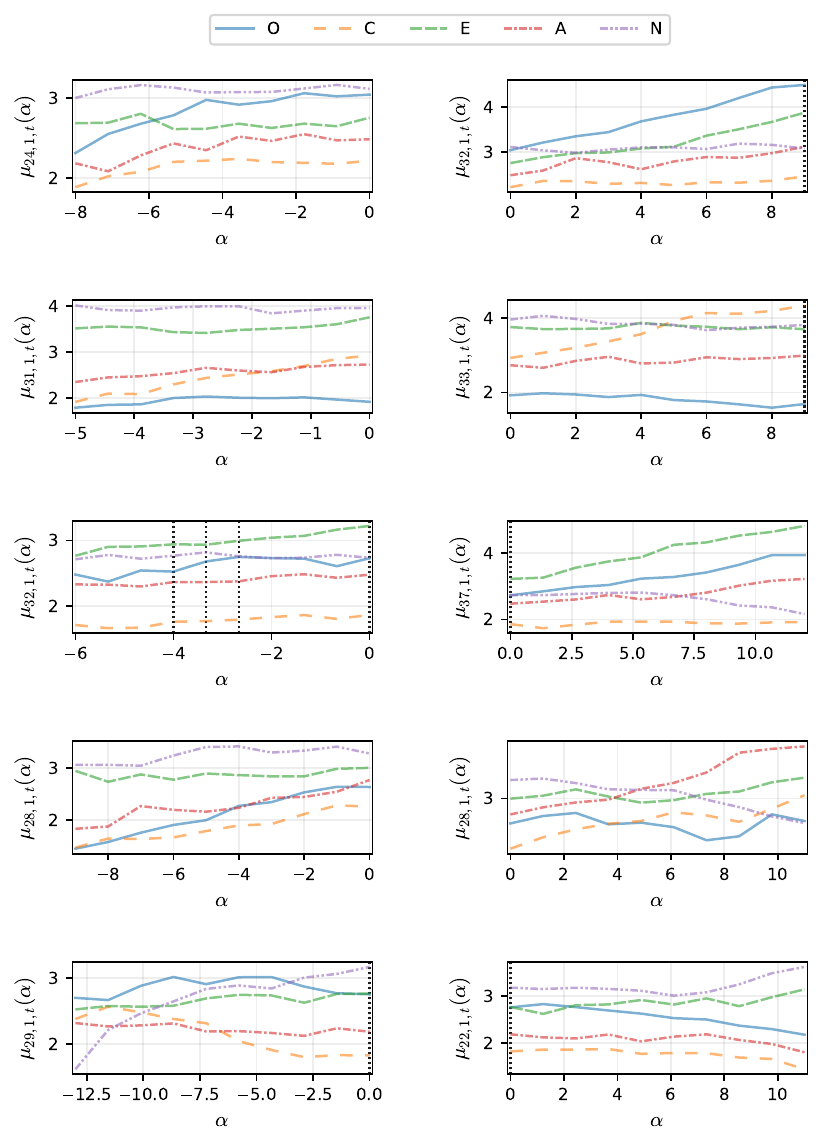}
	\caption{OCEAN scores for gemma-3-27b-it on SJTs, under MDS injections with $s = 1$, using the best-performing layer $\ell$ for each trait-direction pair and 10 equidistant $\alpha$ values from 0 (no steering) to the best-performing $\alpha$.  From top to bottom, rows show openness, conscientiousness, extraversion, agreeableness, and neuroticism results. Negative $\alpha$ steers away from the target construct, and positive $\alpha$ steers toward it. Fluency was evaluated only in the responses to the corresponding SJTs. Vertical lines indicate some nonfluent SJT responses.}
\end{figure}

\clearpage

\section{OCEAN Injection Results for Olmo-3-7B-Instruct}

\begin{figure}[!htbp]
    \centering
    \includegraphics[page=1]{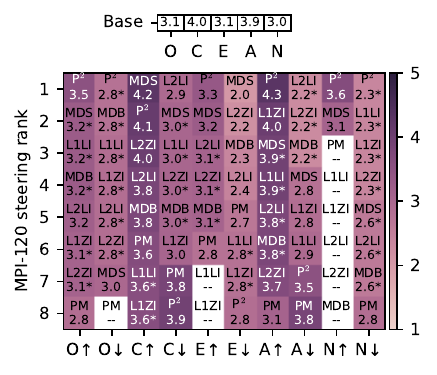}
    \includegraphics[page=2]{analysis/sweeps/Olmo-3-7B-Instruct/heatplots.pdf}
    \caption{Ranking of steering methods on Olmo-3-7B-Instruct by OCEAN trait and direction, and task. Based on each method’s best scores, with asterisks denoting ties in the unrounded results.}
\end{figure}

\begin{figure}[!htbp]
    \centering

    \begin{minipage}[t]{0.49\textwidth}
        \centering
        \includegraphics[page=1]{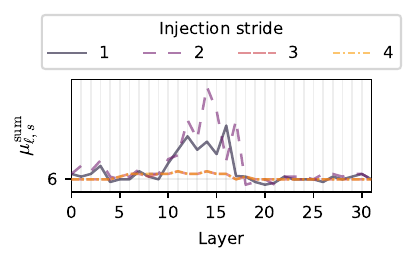}
    \end{minipage}
    \hfill
    \begin{minipage}[t]{0.49\textwidth}
        \centering
        \includegraphics[page=2]{analysis/sweeps/Olmo-3-7B-Instruct/phi_plots.pdf}
    \end{minipage}
    \caption{Overall MDS injections steering performance on Olmo-3-7B-Instruct by injection stride $s$ and model layer $\ell$. The line plot on the left shows MPI-120 results, and the shaded-area plot on the right shows SJT results.}
\end{figure}

\begin{figure}[!htbp]
    \centering
    \includegraphics[page=1]{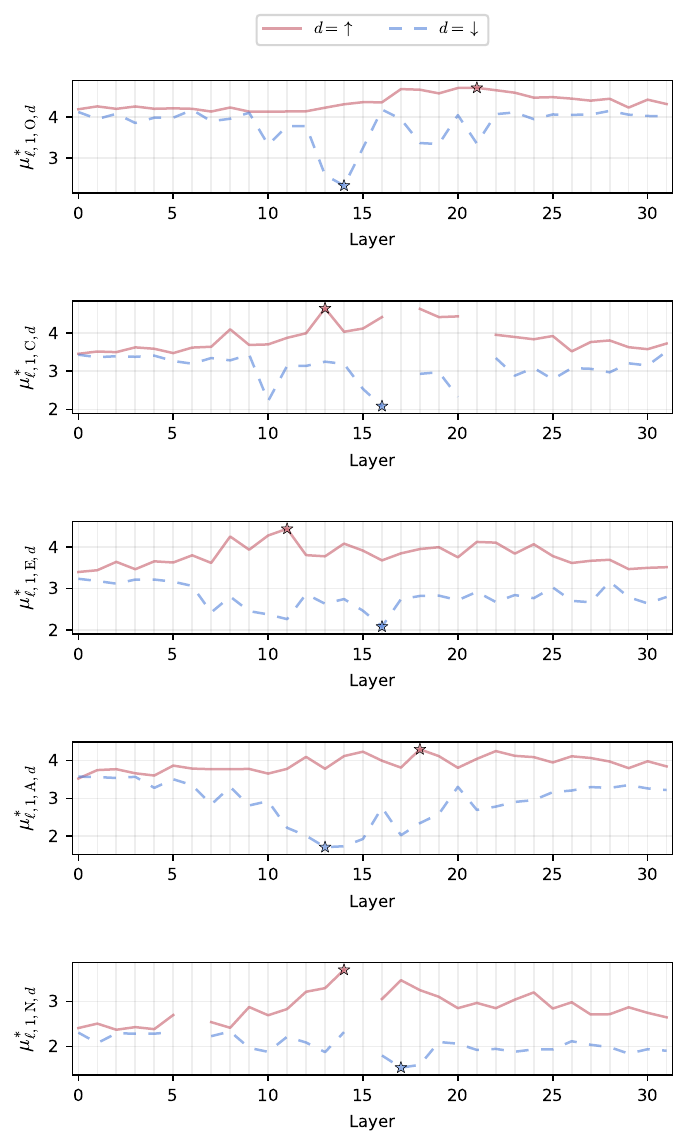}
    \caption{Layerwise extreme OCEAN steering scores on the SJTs task by direction $d \in \{\uparrow, \downarrow\}$ and model layer $\ell$, after applying MDS injections with injection stride $s = 1$ on Olmo-3-7B-Instruct. Stars mark the strongest steering effects across layers ($\phi_{1,t,d}$).}
\end{figure}

\clearpage
\begin{figure}[!htbp]
	\centering
	\includegraphics{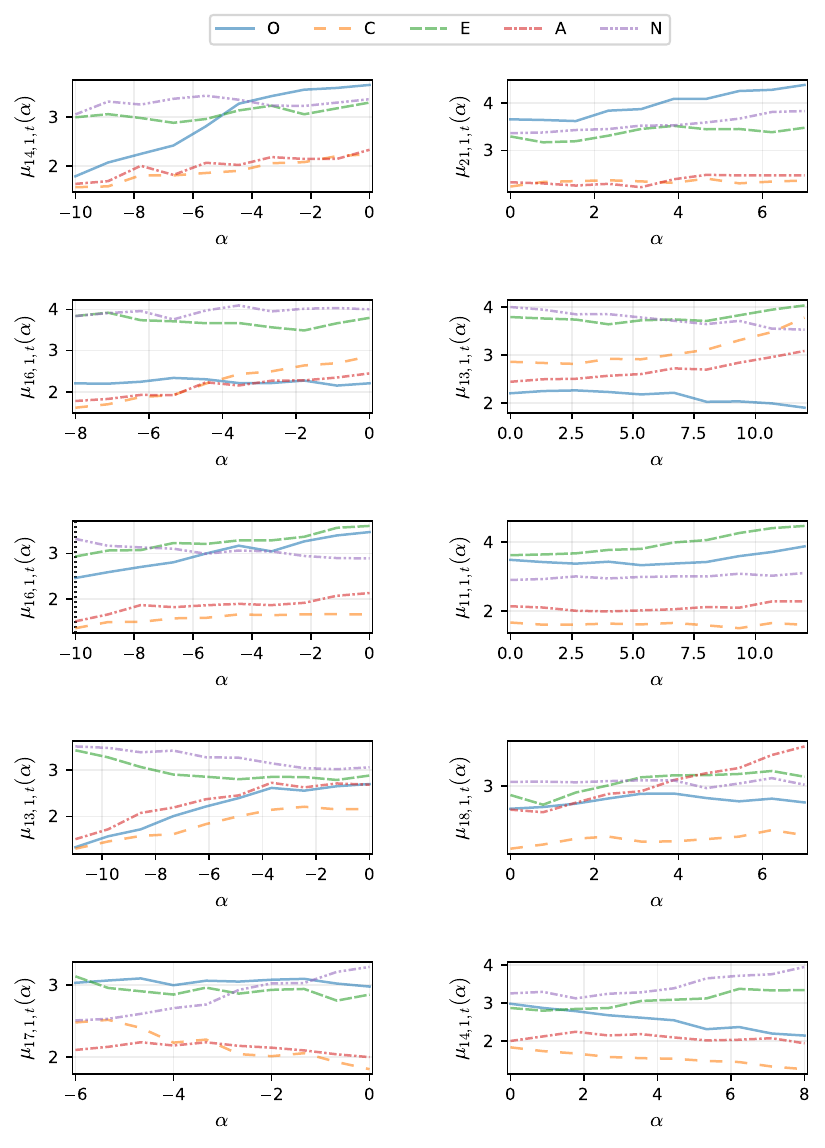}
	\caption{OCEAN scores for Olmo-3-7B-Instruct on SJTs, under MDS injections with $s = 1$, using the best-performing layer $\ell$ for each trait-direction pair and 10 equidistant $\alpha$ values from 0 (no steering) to the best-performing $\alpha$.  From top to bottom, rows show openness, conscientiousness, extraversion, agreeableness, and neuroticism results. Negative $\alpha$ steers away from the target construct, and positive $\alpha$ steers toward it. Fluency was evaluated only in the responses to the corresponding SJTs. Vertical lines indicate some nonfluent SJT responses.}
\end{figure}

\clearpage

\section{OCEAN Injection Results for Olmo-3.1-32B-Instruct}

\begin{figure}[!htbp]
    \centering
    \includegraphics[page=1]{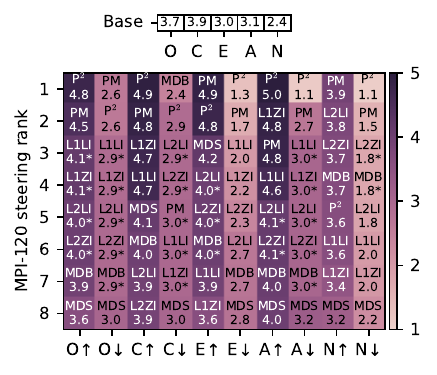}
    \includegraphics[page=2]{analysis/sweeps/Olmo-3.1-32B-Instruct/heatplots.pdf}
    \caption{Ranking of steering methods on Olmo-3.1-32B-Instruct by OCEAN trait and direction, and task. Based on each method’s best scores, with asterisks denoting ties in the unrounded results.}
\end{figure}

\begin{figure}[!htbp]
    \centering

    \begin{minipage}[t]{0.49\textwidth}
        \centering
        \includegraphics[page=1]{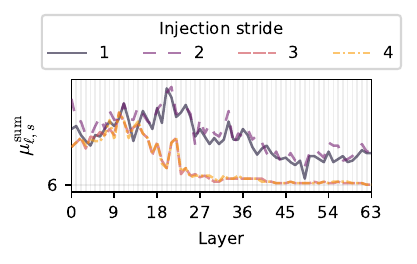}
    \end{minipage}
    \hfill
    \begin{minipage}[t]{0.49\textwidth}
        \centering
        \includegraphics[page=2]{analysis/sweeps/Olmo-3.1-32B-Instruct/phi_plots.pdf}
    \end{minipage}
    \caption{Overall MDS injections steering performance on Olmo-3.1-32B-Instruct by injection stride $s$ and model layer $\ell$. The line plot on the left shows MPI-120 results, and the shaded-area plot on the right shows SJT results.}
\end{figure}

\begin{figure}[!htbp]
    \centering
    \includegraphics[page=1]{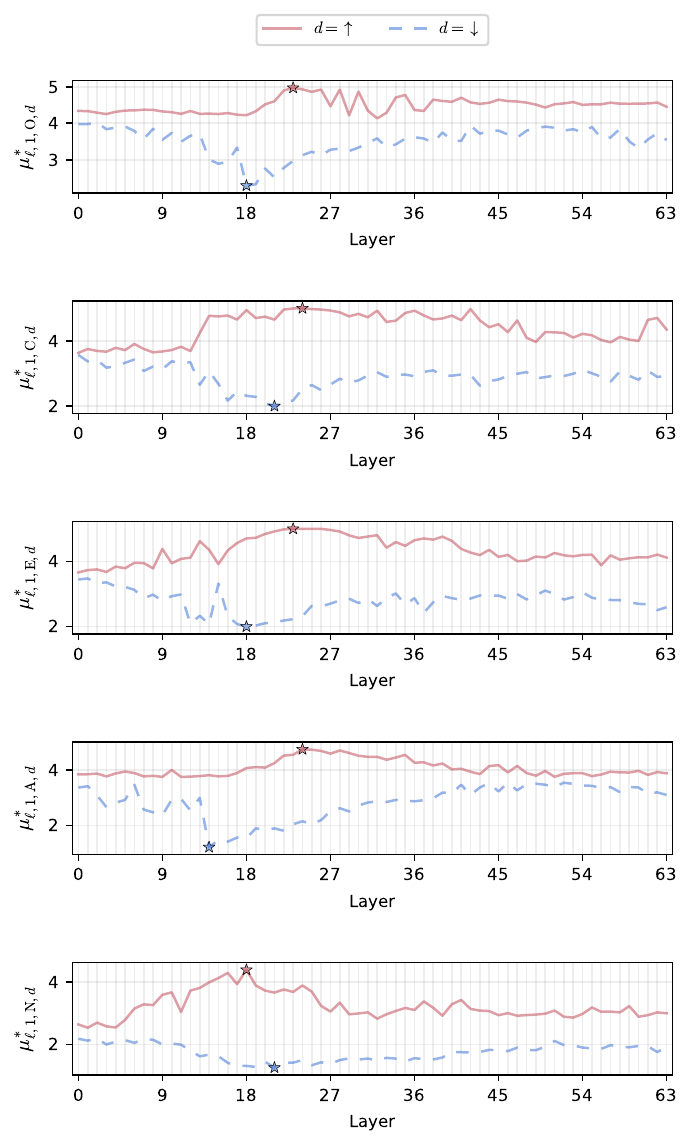}
    \caption{Layerwise extreme OCEAN steering scores on the SJTs task by direction $d \in \{\uparrow, \downarrow\}$ and model layer $\ell$, after applying MDS injections with injection stride $s = 1$ on Olmo-3.1-32B-Instruct. Stars mark the strongest steering effects across layers ($\phi_{1,t,d}$).}
\end{figure}

\clearpage
\begin{figure}[!htbp]
	\centering
	\includegraphics{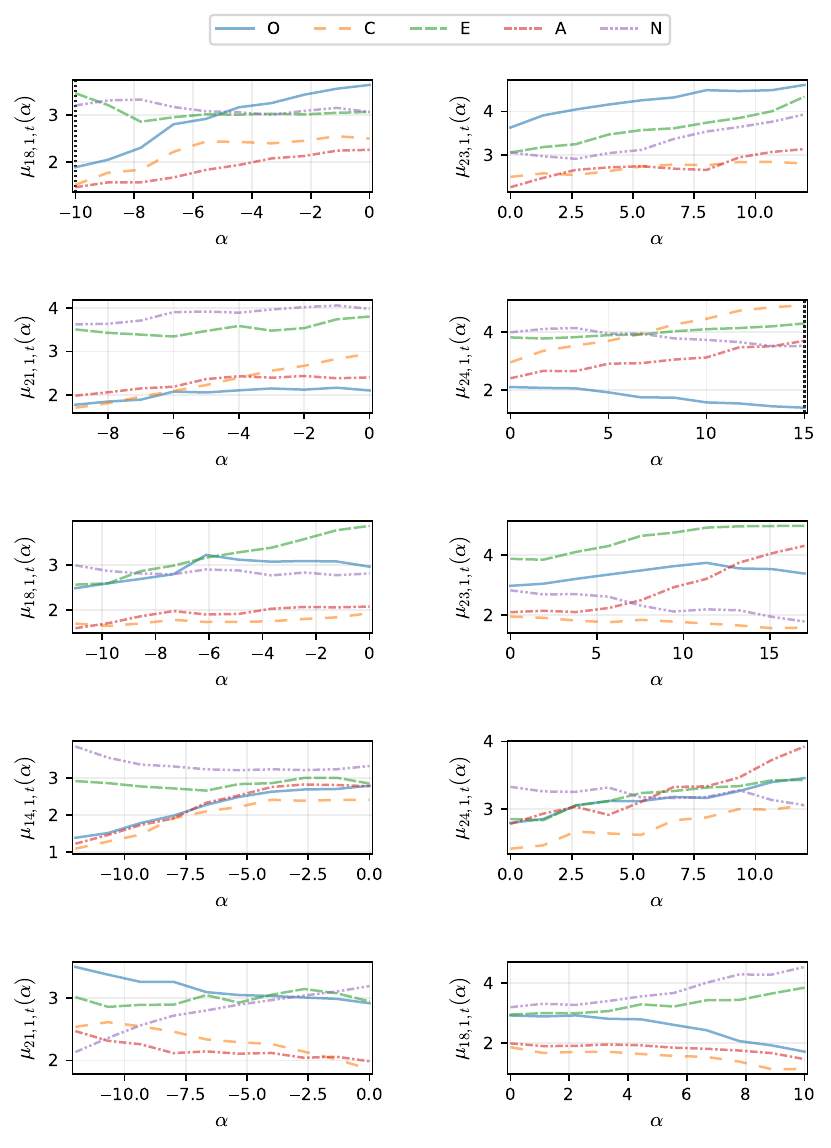}
	\caption{OCEAN scores for Olmo-3.1-32B-Instruct on SJTs, under MDS injections with $s = 1$, using the best-performing layer $\ell$ for each trait-direction pair and 10 equidistant $\alpha$ values from 0 (no steering) to the best-performing $\alpha$.  From top to bottom, rows show openness, conscientiousness, extraversion, agreeableness, and neuroticism results. Negative $\alpha$ steers away from the target construct, and positive $\alpha$ steers toward it. Fluency was evaluated only in the responses to the corresponding SJTs. Vertical lines indicate some nonfluent SJT responses.}
\end{figure}

\end{document}